\documentclass[twoside,11pt]{article}

\usepackage[accepted]{melba}
\usepackage{comment}
\usepackage{multirow}
\graphicspath{ {figures/}}
\usepackage{subcaption}
\usepackage{caption}
\usepackage{graphicx}
\usepackage{soul}
\usepackage[dvipsnames]{xcolor}
\makeatletter
\newcommand{\@chapapp}{\relax}%
\makeatother
\newcommand{\cmark}{\ding{51}}

\usepackage{bbding}
\usepackage{pifont}
\usepackage{wasysym}
\usepackage{amssymb}
\usepackage{booktabs}

\usepackage{mwe} 

\usepackage{amsmath,amsfonts}

\makeatletter
\renewcommand*{\thanks}[1]{%
  \footnotemark
  \protected@xdef\@thanks{\@thanks
    \protect\footnotetext[\arabic{footnote}]{#1}}%
}
\makeatother

\melbaheading{2022:029}{https://www.melba-journal.org/papers/2022:029.html}{2022}{1-37}{07/2022}{12/2022}{Nichyporuk and Cardinell \textit{et al}}{2021 MICCAI Workshop Omnibus Special Issue -- DART Guest}{Ziyue Xu, Konstantinos Kamnitsas}

\ShortHeadings{Impact of Annotation Style in Medical Image Segmentation}{Nichyporuk and Cardinell \it{et al.}}
\firstpageno{1}

\title{Rethinking Generalization: The Impact of Annotation Style on Medical Image Segmentation}

\author{\name{Brennan Nichyporuk\thanks{Contributed Equally}} \email nichypob@mila.quebec \\ 
	\addr MILA  (Quebec  Artificial  Intelligence  Institute),  Montreal, Canada \\ 
	Centre for Intelligent Machines, McGill University, Canada
	\AND
	\name {Jillian Cardinell$^*$} \email jcardine@cim.mcgill.ca \\
    \addr Centre for Intelligent Machines, McGill University, Canada \\ MILA  (Quebec  Artificial  Intelligence  Institute),  Montreal, Canada
    \AND
    \name Justin Szeto \email jszeto@cim.mcgill.ca \\
    \addr Centre for Intelligent Machines, McGill University, Canada \\ MILA  (Quebec  Artificial  Intelligence  Institute),  Montreal, Canada
     \AND
    \name Raghav Mehta \email raghav@cim.mcgill.ca \\
    \addr Centre for Intelligent Machines, McGill University, Canada \\ MILA  (Quebec  Artificial  Intelligence  Institute),  Montreal, Canada
     \AND
    \name Jean-Pierre R. Falet \email jpfalet@cim.mcgill.ca \\
    \addr Department of Neurology and Neurosurgery, McGill University, Canada \\ Centre for Intelligent Machines, McGill University, Canada \\ MILA  (Quebec  Artificial  Intelligence  Institute),  Montreal, Canada
     \AND
	\name Douglas L. Arnold \email douglas.arnold@mcgill.ca \\
    \addr Department of Neurology and Neurosurgery, McGill University, Canada \\ NeuroRx Research, Montreal, Canada
    \AND
	\name Sotirios A. Tsaftaris \email S.Tsaftaris@ed.ac.uk \\
    \addr School of Engineering, University of Edinburgh, UK \\ The Alan Turing Institute, UK
    \AND
    \name Tal Arbel \email arbel@cim.mcgill.ca \\
    \addr Centre for Intelligent Machines, McGill University, Canada \\ MILA  (Quebec  Artificial  Intelligence  Institute),  Montreal, Canada
}

\begin{document}

\maketitle

\begin{abstract}
Generalization is an important attribute of machine learning models, particularly for those that are to be deployed in a medical context, where unreliable predictions can have real world consequences. While the failure of models to generalize across datasets is typically attributed to a mismatch in the data distributions, performance gaps are often a consequence of biases in the ``ground-truth" label annotations. This is particularly important in the context of medical image segmentation of pathological structures (e.g. lesions), where the annotation process is much more subjective, and affected by a number underlying factors, including the annotation protocol, rater education/experience, and clinical aims, among others. In this paper, we show that modeling annotation biases, rather than ignoring them, poses a promising way of accounting for differences in annotation style across datasets. To this end, we propose a generalized conditioning framework to (1) learn and account for different annotation styles across multiple datasets using a single model, (2) identify similar annotation styles across different datasets in order to permit their effective aggregation, and (3) fine-tune a fully trained model to a new annotation style with just a few samples. Next, we present an image-conditioning approach to model annotation styles that correlate with specific image features, potentially enabling detection biases to be more easily identified. 
\end{abstract}


\section{Introduction}
Supervised deep learning medical image segmentation models rely heavily on annotations. However, in the context of segmentation of focal pathologies (e.g. lesions, tumours) from medical images,  obtaining absolute ``ground truth'' annotations is not possible due to limitations in acquisition sequences, partial volume effects~\citep{GONZALEZBALLESTER2002389}, ambiguity in diseased tissue boundaries, among other factors. As a result, annotations are subjective, even when produced by the most skilled expert raters, and result in high \textit{inter-rater} variability. Establishing a fixed annotation process can help reduce inter-rater variability, but also introduces biases that are embedded into the annotation process itself. These biases may include the automated method used to generate the initial annotations (if any) before expert correction, the objectives of the study (e.g. diagnosis, counting lesions, volumetric measurements), or the instructions provided to raters. Since performance metrics are computed with respect to these ``ground truth'' annotations, they must be interpreted with some care, particularly when generalization performance is measured on held-out datasets which may be subject to completely different annotation biases.

\begin{figure}[h]
  \centering
  \begin{subfigure}[h]{\textwidth}
  \centering
      \includegraphics[width=0.65\textwidth]{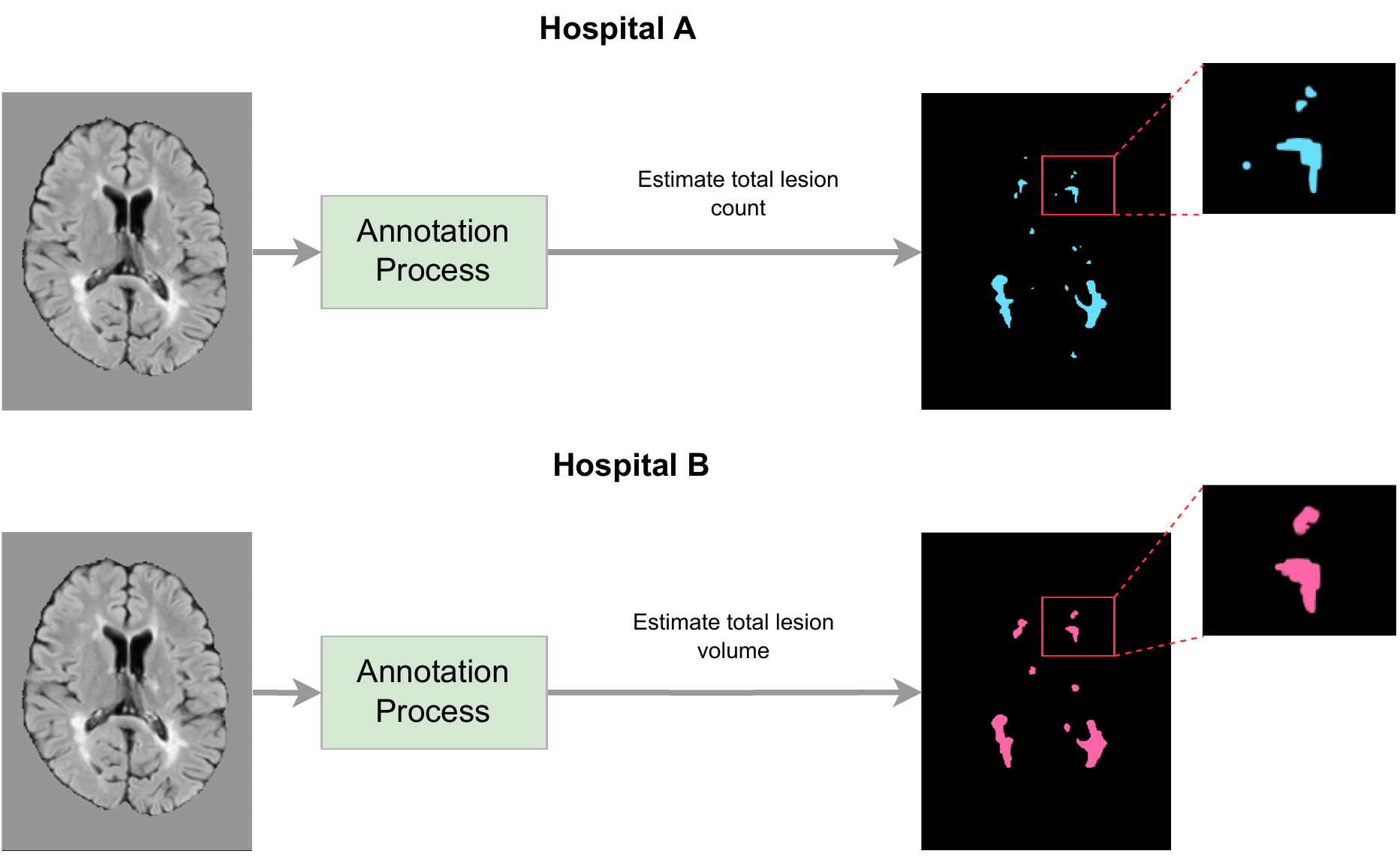}
      \caption{Example case demonstrating how changing the labeling task per dataset can affect the final labels.}
      \label{fig:style_ex}
  \end{subfigure}
  \begin{subfigure}[h]{\textwidth}
  \centering
      \includegraphics[width=0.65\textwidth]{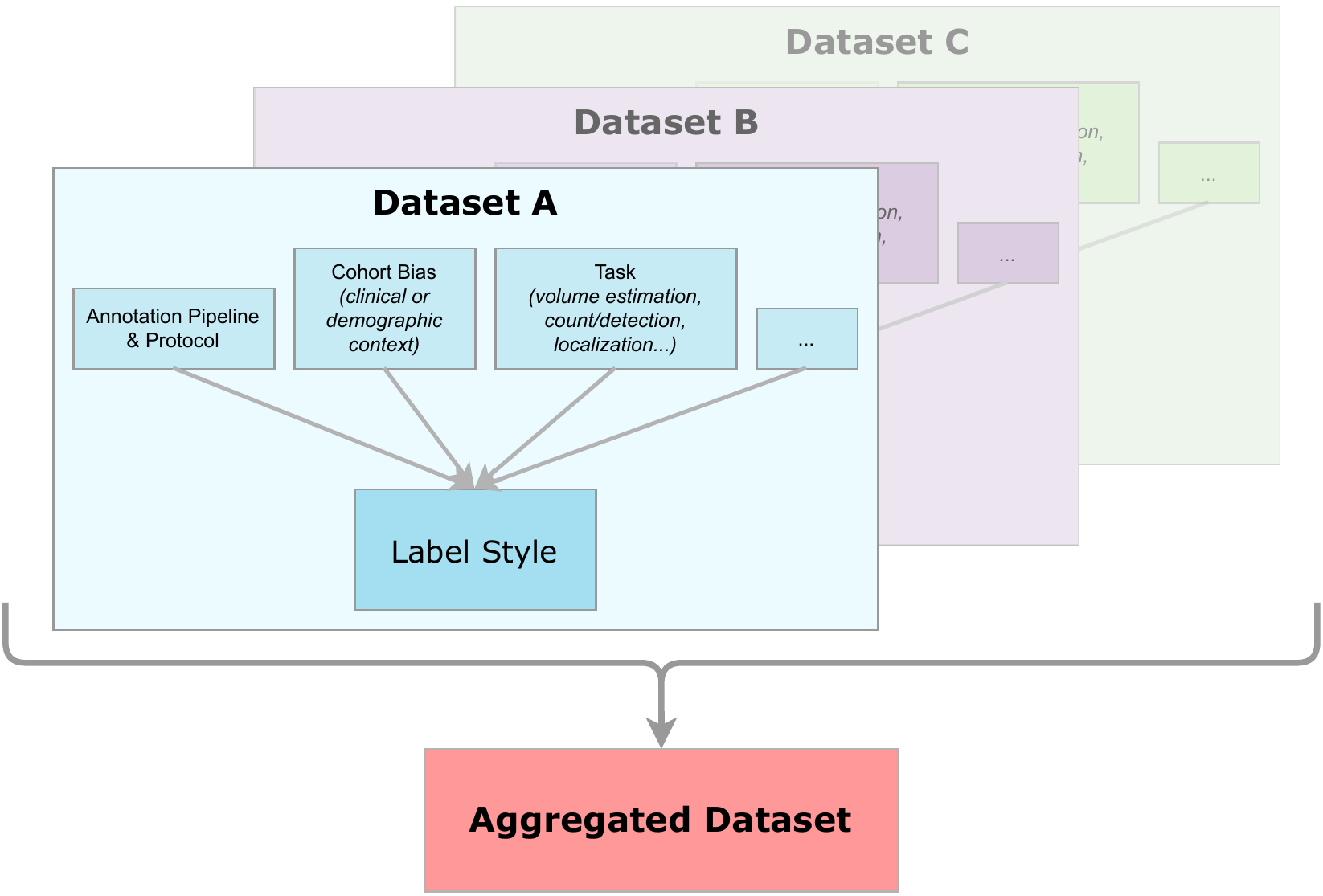}
      \caption{Example depicting several factors contributing to label style in aggregated datasets.}
      \label{fig:agg_dataset}
  \end{subfigure}
  \caption{Illustration of the sources of label bias in aggregated datasets.}
  \label{fig:label_bias}
\end{figure}

To illustrate how annotation biases arise in practice, consider an example where Hospital A is measuring lesion counts in multiple sclerosis (MS) patients, and where Hospital B is measuring lesion volumes in MS patients, as depicted in Figure~\ref{fig:style_ex}. Both tasks involve lesion segmentation, but the resulting ``ground-truth'' annotations are quite different. Hospital A's annotation style results in more, but smaller lesions, while Hospital B's annotation style results in larger and fewer lesions. In this example, a model trained on either dataset would see a performance drop on the other, not necessarily due to a distribution shift in the image-space (e.g. scanner, image protocol, etc.), but due to a distribution shift in the label-space. Although there have been a number of publications addressing distribution shifts in the image-space~\citep{biberacher_intra-_2016, karani_lifelong_2018, van_opbroek_transfer_2015, HeMIS, shen_brain_2019}, this work primarily focuses on the open problem of addressing distribution shifts in the label-space. 

There are many inter-related factors that can have an impact on the annotation style of any given dataset. For instance, many medical image pathology segmentation datasets include semi-manual annotations~\footnote{Semi-manual annotation is a process by which trained experts make corrections to the output of an automatic method.}, where differences in the underlying automated method used, or even the degree of expert correction can have a significant influence on the final annotation. Semi-manual methods are typically used when there is a need to prioritize consistency in the labelling, for example in order to measure changes in lesion load across subsequent scans. Indeed, Hospital B (as depicted in Figure~\ref{fig:style_ex}) may have opted for a semi-manual approach if there was a need to evaluate the effect of treatment on changes in lesion volume over time. On the other hand, Hospital A may have opted for a fully manual approach in order to best characterize the distribution of lesion counts across the population. While both sets of annotations would be entirely valid for the particular downstream clinical task of interest, they would result in  completely different annotation styles. Moreover, several papers have identified annotation biases that result from different annotation styles across raters, even in contexts where labels are derived manually using a consistent annotation protocol~\citep{chotzoglou_exploring_2019, vincent_impact_2021, jungo_effect_2018, shwartzman_worrisome_2019}. Even if all the aforementioned factors could be completely controlled for, it is impossible to completely eliminate detection biases (i.e. biases that result from a systematic difference with respect to how outcomes are measured or assessed across identifiable groups).   

Given that annotation styles may differ across datasets, the \textit{perceived} generalization performance of any method will be affected as a result. Recent domain-invariant methods~\citep{kamnitsas2017unsupervised, lafarge2019learning, ilse2020diva, sun2016deep} are incredibly useful in learning robust feature extraction from the images. However, annotation style differences are impossible for these methods to overcome. Furthermore, given that the labels are dependent on the circumstances of their collection, the final labels used to evaluate the ability of the approach to generalize will inherently be domain-dependent. Due to this complication, being completely invariant to the source domain will ignore information regarding annotation style differences that can affect the final results~\citep{bui2021exploiting, wang2022embracing}. Utilising information relevant to the source domain's annotation style within automated algorithms can therefore help researchers perform informed evaluations and help them obtain the labels desired for their task. This can be seen by revisiting the example scenario with Hospital A and Hospital B once again. A reasonable scenario would be to attempt to pool the two datasets together in order to obtain a larger dataset on which to train an automated algorithm. Being domain invariant means desiring a model independent of the source of the dataset, which may result in labels that do not reflect each institution's respective goals. Therefore, in order to aggregate the two datasets and learn from the additional data while still generating the desired annotation style for both hospitals, meta-information regarding the annotation process will be required. If another institution, Hospital C, had a similar annotation process to Hospital A, identifying this similarity would be a key piece of information, as it would notify researchers that data from Hospital A and Hospital C could potentially be considered a single annotation style. Given that identifying similar annotation styles is not particularly straightforward due to the complexity of the annotation process, data-driven techniques that would enable similar annotation styles to be identified would be very useful. The overall benefit of considering auxiliary information relevant to the annotation style is that the model can more strategically leverage data from multiple sources, while accommodating each source's unique requirements and performing a fair evaluation.

In this work, we examine the relationship between annotation biases and conventional notions of generalization in the context of medical image pathology segmentation. We show that modelling annotation biases, rather than ignoring them, poses a promising way to account for annotation style differences across datasets. To this end, we propose a generalized conditioning framework to (1) learn and account for differences in annotation style across multiple datasets using a single model, (2) identify similar annotation styles across datasets in order to permit their effective aggregation, and (3) fine-tune a fully trained model to a new annotation style with just a few samples. Lastly, we present a novel image-conditioning approach to model annotation styles that correlate with specific image features, potentially enabling detection biases to be more easily identified. We evaluate our approach using six different MS clinical trial datasets and provide an in-depth analysis of the challenges that different annotation styles can present to supervised medical image segmentation approaches, both in terms of aggregating multiple independent datasets together, and in terms of evaluating the generalization performance of segmentation models more broadly. 

This paper extends our previously published paper~\cite{SCIN} with a number of additional contributions: First, we present additional experiments and a more in-depth analysis of the proposed generalized conditioning framework, using a total of six clinical trial datasets compared to the three clinical trial datasets used in our original publication. Second, we propose a new similarity analysis method to identify similar annotation styles across datasets and demonstrate how it could be used in practice. Third, we propose an image conditioning approach to model annotation styles that correlate with specific image features, potentially enabling detection biases to be more easily identified. 

\section{Background and Related Work}
In this section, we summarize related work detailing how annotation biases arise, and review research aiming to correct or account for these biases. We then briefly touch upon related work on adaptive normalisation methods.

\noindent\textbf{Sources of Annotation Biases in Aggregated Datasets}
Other researchers have noticed challenges in aggregating different datasets for automatic methods for a variety of reasons. Most research is geared at identifying and addressing variations across human raters. \cite{liao2021modeling} showed that inter-rater biases can be problematic for automated methods. They noted that stochastic rater errors are more easily solved, and that consistent rater biases are the primary issue. This finding was also presented by~\cite{landman2012foibles}, where they concluded that for collaborative labelling efforts or datasets to be successful, unbiased labellers are required. However, the subjectivity involved in attempting to model "ground truth" in medical pathology segmentation tasks result in substantial rater variability \citep{cross-domain-xray}. For instance, \cite{altay2013reliability} have noted that lesion segmentation in multiple sclerosis (MS), especially new and enlarging lesion segmentation, can result in high variability among expert raters. The fact that the border between MS lesions and their surrounding healthy tissues represents a continuous (rather than discrete) transition in voxel intensity, coupled with inherent inconsistencies within a given lesion's intensity profile, makes it particularly challenging for different human raters to reach the same conclusion as to the exact border location from visual inspection alone. Depending on the display contrast when a rater is viewing an MRI for manual segmentation, the lesion size may appear slightly larger or smaller than for another rater who might be viewing under different display conditions. The ambiguity of MS lesion borders is further amplified by the existence of an intermediate, pre-lesional abnormality called diffusely abnormal white matter (DAWM)~\citep{dadar2021diffusely}. Because there is no universally accepted definition for DAWM, some raters might include more DAWM within their lesion masks than others, leading to arbitrarily different annotation styles. Lastly, detection and observer bias are an important source of inter-rater variability that can be implicit and therefore poses a challenge in terms of explainability. \cite{hrobjartsson2013observer} conducted a study across many different clinical trials and found a significant difference in labels from raters that were fully blinded (to the intervention), compared to non-blinded raters. Blinding is often thought of as useful for eliminating confounding factors. However, in the case of segmentation, \cite{freeman2021iterative} remarked that some annotators that are blinded to medical history may produce potentially "unacceptable" annotations as compared to raters with access to full medical records. These factors can lead to cohort biases~\footnote{A cohort is a group of people who share a common defining characteristic.}, in which the annotations are influenced by demographic or clinical information of the cohort being labelled.

In addition to human factors, the label generation process itself can be a major contributor to annotation style~\citep{shinohara2017volumetric}. \cite{freeman2021iterative} noted that the guidelines provided to raters had an impact on quality and annotation outcome. \cite{vrenken2021opportunities} also found that differences in labelling protocols can degrade algorithm performance, as well as limit to what extent performance can be validated. \cite{oakden2020exploring} recommend that labelling rules and other data generation procedures be made available to users due to their impact on testing results. \cite{cross-domain-xray} further discuss the impact of the use of semi-automated labelling software in the label generation process and resulting labels. The problem of differences in the annotation process has even been noted in natural imaging tasks, which require less expertise and are significantly less subjective \citep{recht2019imagenet}. Overall, many factors combine to influence the annotation style across different datasets. Although many researchers have identified these problems, most attempts at addressing and/or accounting for sources of annotation style are primarily focused on situations where individual biases can be isolated, such as when a single sample is given to two or more raters to annotate. However, there are no studies that we are aware of that directly address the much more complex biases that can arise across datasets as a consequence of the label generation process as a whole in medical image analysis (e.g. goal of the study, rater experience/education, instructions provided to raters, etc.).

\noindent\textbf{Accounting for Inter-Rater Bias}
As previously mentioned, several studies have focused on addressing the impact of inter-rater biases on the resulting annotation style. These studies primarily use datasets where multiple raters are each given the same dataset to segment. In these cases, differences in labels can be directly associated with the raters' opinion or style, experience level, or the uncertainty of the target pathology~\citep{heller_imperfect_2018, warfield_simultaneous_2004, joskowicz_inter-observer_2019, jungo_effect_2018, vincent_impact_2021}. Other researchers refer to the problem of inter-rater bias as a source of ``label noise" \citep{ji_learning_2021, zhang_disentangling_2020, karimi_deep_2020}. These approaches operate on the assumption that the variability in labels can be attributed to rater biases, and that differences between raters reflect a noisy variation around the real ``ground truth". While this assumption may be adequate for the segmentation of healthy tissues, ``ground truth" for pathology segmentation tasks can be incredibly arbitrary, with multiple possible variants of ground truth considered equally valid. In contrast, we deal with the task of aggregating multiple datasets where the entire labelling process may differ, and where only a \textit{single annotation} (from one of multiple possible raters) is given for each sample, making individual biases difficult to disentangle. Indeed, in this paper, each dataset has its own individual annotation style that results from the label generation process of each dataset.

\noindent\textbf{Adaptive Normalization-Based Methods}
For completeness, we briefly touch on the use of adaptive normalization methods, which have proven useful in previous bias adaptation and style transfer tasks~\citep{D-BIN, batch-instance-norm, karani_lifelong_2018, cartoonGAN, Kim2020TransferLF, Ruta2021ALADINAL,jacenkow2020inside}. Several papers focus on natural imaging problems, such as artistic style transfer and image denoising~\citep{D-BIN,batch-instance-norm, cartoonGAN,Kim2020TransferLF, Ruta2021ALADINAL}. Researchers have also applied conditional normalization methods to model imaging biases~\citep{karani_lifelong_2018}, or to leverage relevant clinical information~\citep{jacenkow2020inside}. 

\section{Methodology}
\begin{figure}[h]
\centering
  \centering
   \includegraphics[width=0.6\textwidth]{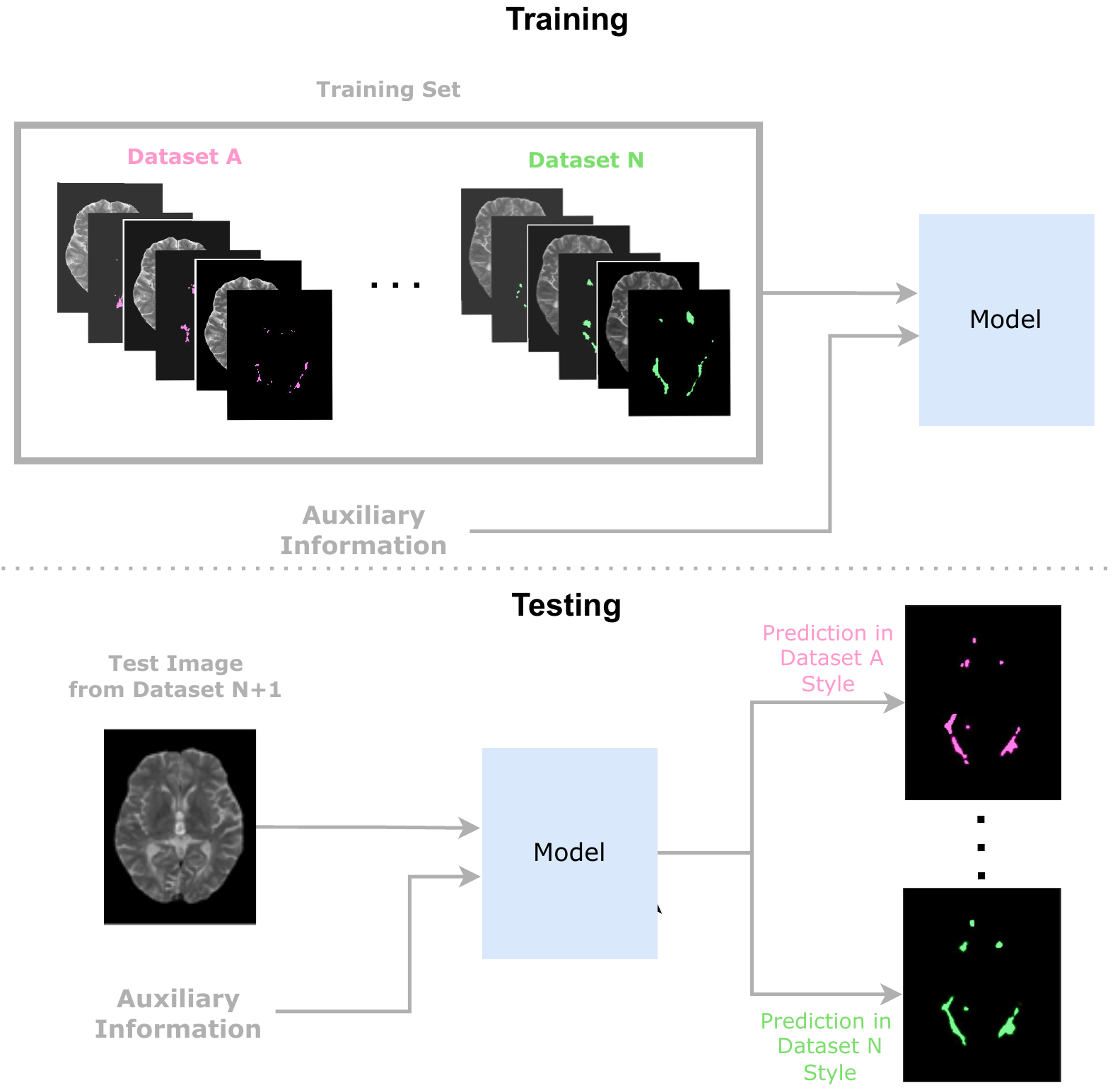}
  \caption{System overview. (Top) Training module: Training on multiple datasets with auxiliary information to learn the associated bias for each dataset. (Bottom) Testing module: Auxiliary information used to generate multiple lesion segmentation maps, each with a different annotation style, for the input test image.}
  \label{fig:flow}
\end{figure}
In this paper, we propose using Conditioned Instance Normalization (CIN) to pool multiple datasets while learning each dataset's associated annotation style~\citep{SCIN}. To do this, we provide the input images alongside auxiliary information to condition on, which allows the model to learn the associated annotation styles. An overview of the general approach is shown in Figure~\ref{fig:flow}.

\begin{figure}[h]
\centering
  \centering
   \includegraphics[width=\textwidth]{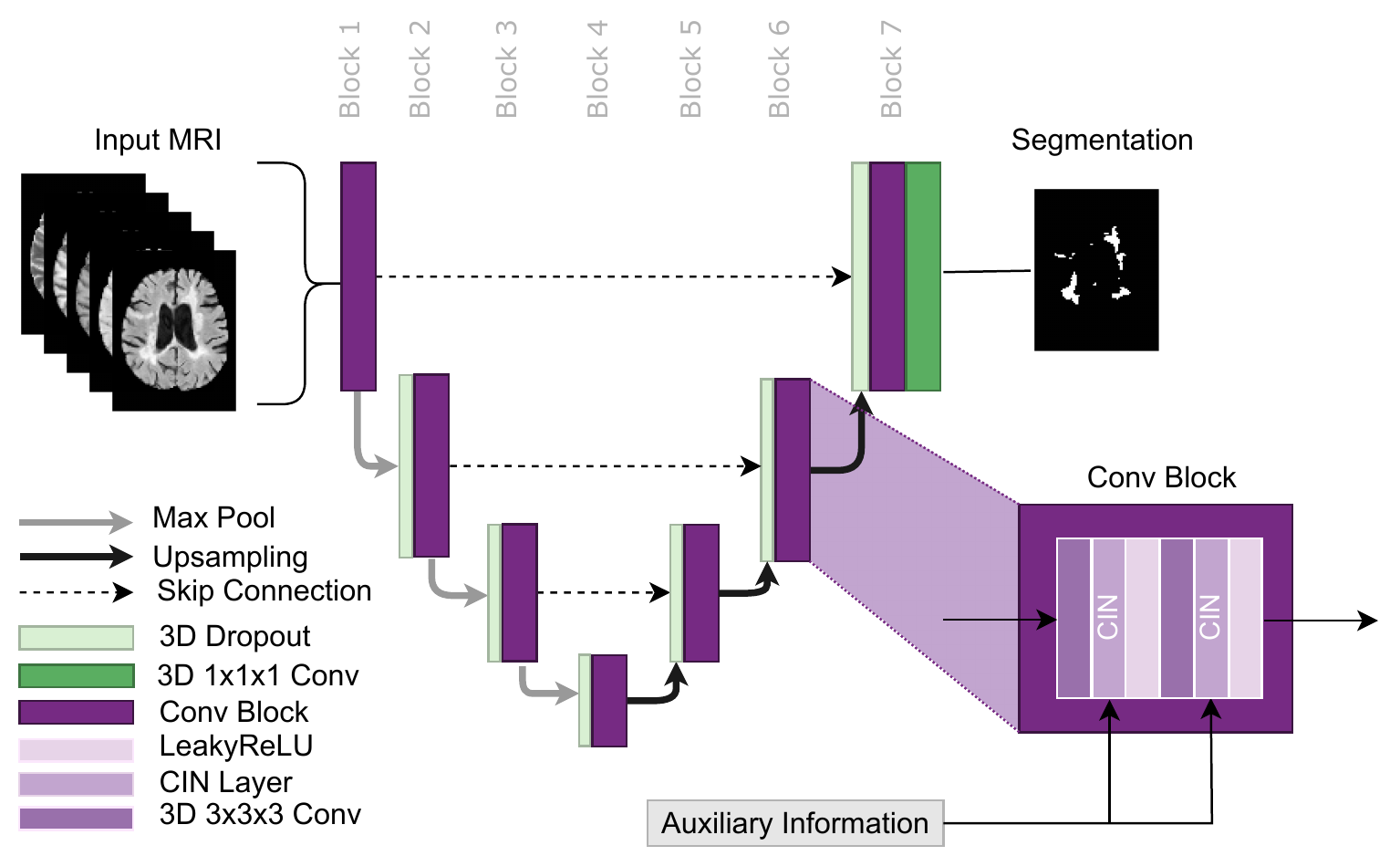}
  \caption{Left: Overview of modified nnUNet (\cite{nnUnet}) architecture used to segment MS T2 lesions. Right: Detail of a conv block. It consists of a series of 3D 3x3x3 Convolution Layer, CIN layer, and a LeakyReLU activation layer. }
  \label{fig:architecture}
\end{figure}

The approach utilises the conditional instance normalization (CIN) layer proposed by \cite{CIN} to learn dataset-specific biases using a set of scale and shift parameters unique to each dataset. This layer allows for dataset biases to be modelled while simultaneously performing the target task. The auxiliary information was provided to the model in the form of a categorical variable, which identifies which set of scale and shift parameters to use for a given input image. In this paper, we focus on modelling dataset biases that lead to unique annotation styles. The approach can potentially be used in any network architecture where Instance Normalization is traditionally applied. Thus the approach is flexible and could potentially be applied for any target task, such as detection, classification, or regression. In this paper, we use CIN in an nn-UNet based architecture for MS lesion segmentation as shown in Figure~\ref{fig:architecture}.

The approach works by scaling and shifting the normalized activations of each layer using a dataset-specific set of affine parameters during the forward pass. The CIN layer is represented by the following equation:

$$\textbf{\text{CIN}} (\mathbf{z}) = \boldsymbol{\gamma}_s \left(\frac{\mathbf{z}-\boldsymbol{\mu}(\mathbf{z})}{\boldsymbol{\sigma}(\mathbf{z})}\right) + \boldsymbol{\beta}_s$$
where $\boldsymbol{\gamma}_s$ and $\boldsymbol{\beta}_s$ are the conditional affine parameters and where $\mu(\mathbf{z})$ and $\sigma(\mathbf{z})$ represent the per-channel mean and standard deviation, respectively. The affine parameters correspond to a specific dataset, $s$. The dataset-specific affine parameters are learned only from samples from the corresponding dataset. The dataset-specific scale and shift parameters are all initialized at 1 and 0 respectively. This initialization ensures that the model learns to apply scale and shift as needed, from a no-scale and no-shift starting point. Other than the dataset-specific affine parameters in each CIN layer, all other network parameters are learned from all samples regardless of dataset identity. This allows the approach to leverage multiple datasets to learn common features while still taking into account dataset-specific annotation styles. A full system overview can be found in Figure~\ref{fig:flow}.

\subsection{Subgroup Discovery with CIN Parameter Analysis}
\label{cin_method}
We now explore how the learned dataset-specific scale and shift parameters of the fully-trained network can be used to identify relationships between annotation styles across different datasets. Understanding these relationships would not only improve data collection methods and make for more fair performance evaluations, but would enable strategic pooling of compatible styles. This type of strategic pooling is especially important for research projects with limited datasets where each style may only have a few labelled samples.

To identify relationships, we calculate the cosine similarity (normalized dot product) of the parameters between all datasets for all CIN layers in the network in order to determine the relevant relationships and where they occur within the network. Specifically, the cosine similarity between two sources, $s$ and $\tilde{s}$, is computed as follows: 
$$\text{Cosine Similarity}_{scale_{s,\tilde{s}}} (n) = \frac{(\boldsymbol{\gamma}_{n_s}-\mathbf{1}) \cdot (\boldsymbol{\gamma}_{n_{\tilde{s}}}-\mathbf{1})}{|\boldsymbol{\gamma}_{n_s}-\mathbf{1}||\boldsymbol{\gamma}_{n_{\tilde{s}}}-\mathbf{1}|}$$ 

$$\text{Cosine Similarity}_{shift_{s,\tilde{s}}} (n) = \frac{\boldsymbol{\beta}_{n_s} \cdot \boldsymbol{\beta}_{n_{\tilde{s}}}}{|\boldsymbol{\beta}_{n_s}||\boldsymbol{\beta}_{n_{\tilde{s}}}|} $$
where $n$ is the specific CIN layer in the network, and $s$ and $\tilde{s}$ represent different sources \footnote{Since the scale parameters are initialized at 1, and a scale of 1 is representative of the parameter having no effect on the activation, we subtract 1 from all scale vectors before performing the cosine similarity calculation. This effectively relocates the origin to the initialization point of all scale parameters allowing for a more fair evaluation of the parameter changes that occurred during training.}. 

By analysing the cosine similarity between dataset-specific parameters, we can quantify high dimensional direction-based relationships between the scale and shift parameter vectors of the different datasets. An additional analysis, that considers magnitude, is performed by comparing the euclidean norm of these vectors. We attempt to look at the norms to detect any noticeable clusters or trends that may go undetected in the cosine similarity analysis.

\subsection{Modeling Synthetic Detection Biases Using Image Conditioning}
\label{sec:imgcon_method}
Beyond modeling dataset-specific biases, we propose an image conditioning mechanism to model image-related detection biases. To do so, we propose an image conditioning mechanism based on FiLM~\cite{perez_film_2018} consisting of two key components: (1) An image encoder to generate a latent representation of the image; (2) A linear layer for each CIN layer to transform the latent representation of the image into a set of affine instance normalization parameters (per channel scale and shift). By doing this, the model is able to adapt the style of the output segmentation based on (global) image characteristic(s). The full architecture can be found in Figure ~\ref{fig:architecture_imcond}.

\section{Implementation Details}
\subsection{Network Architecture and Training Parameters}
We train a modified nnUNet~\citep{nnUnet} for the task of multiple sclerosis (MS) T2-weighted lesion segmentation. The primary modifications include the addition of dropout, and replacing the standard Instance Normalization layer in each convolutional block with a CIN layer. This architecture is shown in detail in Figure~\ref{fig:architecture}. All models in this paper are trained using Binary Cross Entropy (BCE) Loss and Adam optimizer \citep{kingma2014adam}. Random affine and random contrast augmentations are also used to prevent over fitting. 

\subsection{Data Sources}
In this work, we leverage a large, proprietary dataset consisting of images acquired from MS patients from six different multi-centre, multi-scanner clinical trials. In this context, each trial serves as a different dataset. Each trial contains multi-modal MR sequences of patients with one of three different MS subtypes: Secondary-Progressive (SPMS), Primary-Progressive (PPMS), or Relapsing Remitting (RRMS). RRMS patients suffer from recurrent attacks of neurological dysfunction, usually followed by at least partial recovery~\citep{hurwitz2009diagnosis}. SPMS may eventually develop in patients with RRMS. SPMS is characterized by progressive clinical disability in the absence of relapses. The transition from RRMS to SPMS is gradual and not clearly defined. PPMS is characterized progression in the absence of relapses from the onset of the disease, i.e., without a relapsing remitting stage. SPMS and PPMS subjects are about 10 years older than RRMS patients on average. SPMS patients have larger lesion volumes than RRMS and PPMS patients~\citep{lublin2014defining}.

For all trials, each patient sample consists of 5 MR sequences acquired at 1mm $\times$ 1mm $\times$ 3mm resolution: T1-weighted, T1-weighted with gadolinium contrast, T2-weighted, Fluid Attenuated Inverse Recovery (FLAIR), and Proton Density. In each trial, images were collected across multiple different sites, and multiple different scanners ({\raise.17ex\hbox{$\scriptstyle\sim$}}80 per trial), therefore there were no consistent acquisition differences between trials. T2 lesion labels were generated at the end of each clinical trial, and were produced by an external company, where trained expert annotators manually corrected the output of a proprietary automated segmentation method. Since the trials were completed at different times, labels may have been generated with different versions of the automated segmentation method. Although different expert raters corrected the labels, there was overlap between raters across trials and all raters were trained to follow a similar labelling protocol. Each image has one associated label (i.e. one label style, unlike in inter-rater bias studies). Furthermore, within one trial, the same labelling process was followed in order to keep the labels within the trial consistent. All imaging data was re-processed using a single pre-processing pipeline, resulting in nearly indistinguishable intensity distributions across trials. While image-based biases were effectively normalized out after re-processing, the image effects would have influenced the labels during the original labelling process through the semi-automated algorithm.  In this study, the need to model trial-specific biases stems primarily from subtle differences that accumulate throughout the labelling process. How these differences translate into the resulting annotation styles has not been extensively studied, despite the impact this can have on research into related problems (e.g. domain adaptation).

\subsection{Evaluation Metrics}
Automatic T2 lesion segmentation methods are evaluated with voxel-level segmentation metrics. To evaluate performance, we use DICE (Segmentation F1 score) (\cite{F1}).\footnote{Operating point (threshold) is chosen based on the best segmentation DICE.} We also present Precision Recall Area Under Curve (PR-AUC) for all experiments in the Appendix to provide results independent of threshold selection. For relevant experiments, detection F1 is also provided. Detection F1 is obtained by performing a connected component analysis on the segmentation mask to group lesion voxels in an 18-connected neighbourhood to identify individual lesions~\cite{BUNet}. 

\section{Experiments and Results}
\label{sec:expts}
Several sets of experiments were performed to demonstrate the issue of annotation style and the ability of CIN to model it. In the first set of experiments, a trial-conditioned CIN model is compared to naive-pooling and single-trial baselines to show the importance of considering and accounting for annotation style. A detailed analysis of the trial-specific CIN parameters from the trained model is then performed in order to explore patterns or groupings between the learned annotation styles. Another series of conditioning experiments are performed in order to confirm the power of grouping based on the discovered subgroups. Next, we fine-tune CIN parameters on a held-out trial to show how CIN can be used to quickly adapt an existing model to a new annotation style. We then test the proposed framework on a single-trial dataset where we induce a synthetic bias in half of the samples to study the ability of CIN to model annotation styles that are impossible to infer from the image. Lastly, we propose an image-conditioning mechanism to model annotation styles that correlate with specific image features, potentially enabling detection biases to be more easily identified. 

\subsection{Learning Annotation Styles}
\label{sec:trial_cond}


\begin{table}[h]
\caption{DICE performance on test sets for the Trial-Conditioned model, Naive-Pooling model, and Single-Trial models.}
\begin{subtable}[h]{\textwidth}
\caption {Performance on test sets: experiments on Single-trial models.}
\label{tab:singletrial-f1}
\resizebox{\textwidth}{!}{%
\begin{tabular}{|l|ll|cccccc|}
\hline
\multirow{2}{*}{\textbf{ }} &
\multirow{2}{*}{\textbf{Model}} &
  \multirow{2}{*}{\textbf{\begin{tabular}[c]{@{}l@{}}Training \\ Set\end{tabular}}} &
  \multicolumn{6}{c|}{\textbf{Test Performance (DICE)}} \\ 
& & &\textbf{RRMS-A} &\textbf{RRMS-B} &\textbf{SPMS-A} &\textbf{SPMS-B} &\textbf{PPMS-A} &\textbf{PPMS-B} \\ \hline
\textbf{1} &\textbf{Single-Trial} &RRMS-A &\textbf{0.784} &0.779 &0.776 &0.681 &0.647 &0.646 \\
\textbf{2} &\textbf{Single-Trial} &RRMS-B &0.775 &0.779 &0.766 &0.678 &0.638 &0.646 \\
\textbf{3} &\textbf{Single-Trial} &SPMS-A &0.778 &\textbf{0.785} &\textbf{0.782} &0.679 &0.645 &0.647 \\
\textbf{4} &\textbf{Single-Trial} &SPMS-B &0.691 &0.689 &0.686 &\textbf{0.730} &\textbf{0.709} &0.693 \\
\textbf{5} &\textbf{Single-Trial} &PPMS-A &0.697 &0.699 &0.690 &\textbf{0.731} &0.699 &0.681 \\
\textbf{6} &\textbf{Single-Trial} &PPMS-B &0.669 &0.671 &0.663 &0.682 &0.646 &\textbf{0.742} \\ \hline
\end{tabular}}
\end{subtable}

\begin{subtable}[h]{\textwidth}
\centering
\vspace{10pt}
\caption{Performance on test sets: experiments on Single-Trial, Naive-Pooling, Trial-Conditioned models.}
\label{tab:cond_v_baselines_f1}
\resizebox{\textwidth}{!}{%
\begin{tabular}{|l|llcccccc|}
\hline
\multirow{2}{*}{\textbf{ }} &
\multirow{2}{*}{\textbf{Model}} &
  \multirow{2}{*}{\textbf{\begin{tabular}[c]{@{}l@{}}Training \\ Set\end{tabular}}} &
  \multicolumn{6}{c|}{\textbf{Testing Set (DICE)}} \\
& & &\textbf{RRMS-A} &\textbf{RRMS-B} &\textbf{SPMS-A} &\textbf{SPMS-B} &\textbf{PPMS-A} &\textbf{PPMS-B} \\\hline
\textbf{1} &\textbf{Single-Trial} &RRMS-A &0.784 &- &- &- &- &- \\
\textbf{2} &\textbf{Single-Trial} &RRMS-B &- &0.779 &- &- &- &- \\
\textbf{3} &\textbf{Single-Trial} &SPMS-A &- &- &0.782 &- &- &- \\
\textbf{4} &\textbf{Single-Trial} &SPMS-B &- &- &- &0.730 &- &- \\
\textbf{5} &\textbf{Single-Trial} &PPMS-A &- &- &- &- &0.699 &- \\
\textbf{6} &\textbf{Single-Trial} &PPMS-B &- &- &- &- &- &0.742 \\  \hline
\textbf{7} &\textbf{Naive-Pooling} &\textbf{All} &0.766 &0.754 &0.756 &0.722 &0.700 &0.738 \\
\textbf{8} &\textbf{Trial-Conditioned} &\textbf{All} &\textbf{0.787} &\textbf{0.789} &\textbf{0.788} &\textbf{0.737} &\textbf{0.709} &\textbf{0.744} \\ \hline
\end{tabular}}
\end{subtable}
\end{table}

In this section, experiments are conducted to examine the CIN model and its benefits in situations with variable annotation styles. This series of experiments is conducted with the described 6 clinical trial datasets. To maintain balance, an equal number of patients were taken from each clinical trial, resulting in 390 patients per trial, totalling 2340 patients in the aggregated dataset. This dataset was then then divided (on a per-trial basis) into non-overlapping training (60\%), validation (20\%), and testing (20\%) sets. In these experiments, the CIN model is conditioned on the trial, termed the \textit{trial-conditioned} model. We also train single-trial models where only one dataset is used in training, and a naive-pooling model where all datasets are used but the model is blind to the source of the data. Each model is individually hyperparameter tuned to optimize performance on the validation sets, thus the naive-pooling, single-trial, and conditioned models are all tuned separately to enable fair comparisons.

Table~\ref{tab:singletrial-f1} demonstrates the problem of different annotation styles learned by single-trial models (see Appendix A for more results). These models show a clear performance degradation of up to 10\% when applied to different trial test sets. This explicitly shows the issue of comparing models trained on one annotation style to ground truths generated in a conflicting annotation style. Since we only have one annotation style per trial, it is essentially impossible to accurately and fairly evaluate a single-trial model on another trial with a different annotation style.

\begin{figure}[h]
\centering
  \centering
   \includegraphics[width=\textwidth]{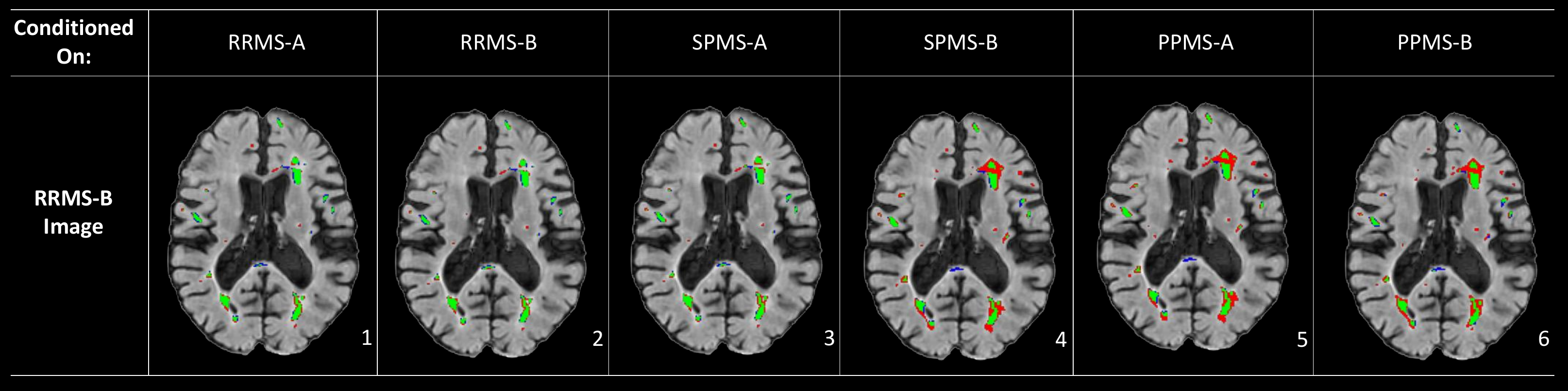}
  \caption{RRMS-B FLAIR example image segmented with the Trial-Conditioned model using different conditioning styles. Green is true positive, red is false positive, and blue is false negative with respect to the RRMS-B ``ground truth" label.}
  \label{fig:define_example}
\end{figure}

Comparisons of the DICE segmentation scores for the single-trial models, naive-pooling model, and the proposed trial-conditioned model are shown in Table~\ref{tab:cond_v_baselines_f1}. The lower overall performance of the naive-pooling model (as is common practice) demonstrates another consequence of variable annotation styles. Specifically, despite a 500\% increase in the size of the training set, the naive-pooling model under performs the single trial baselines in nearly all cases, effectively eliminating the expected benefit of using more data. This is likely because the naive-pooling model has no information regarding the annotation style that is expected for any given sample and when tested on each trial, is unable produce a label in the required style.. On the other hand, provided with context on the source of each patient sample, the trial-conditioned model shows performance on par with the single-trial models, illustrating CIN's ability to learn trial-specific annotation styles. The final trial-conditioned model is also capable of producing any annotation style for any input, thus providing researchers with multiple ``opinions" on a given sample, as shown in Figure \ref{fig:define_example}.

\begin{table}[h]
\caption{DICE performance on test sets for the Trial-Conditioned model with different conditioning styles.}
\label{tab:trialcond-f1}
\resizebox{\textwidth}{!}{%
\begin{tabular}{|l|l|cccccc|}
\hline
\multirow{2}{*}{ } &
\multirow{2}{*}{\textbf{\begin{tabular}[c]{@{}l@{}}Conditioning \\ Style\end{tabular}}} & \multicolumn{6}{c|}{\textbf{Testing Set (DICE)}}      \\
                                       & &\textbf{RRMS-A} &\textbf{RRMS-B} &\textbf{SPMS-A} &\textbf{SPMS-B} &\textbf{PPMS-A} &\textbf{PPMS-B} \\\hline
\textbf{1} &RRMS-A &\textbf{0.787} &\textbf{0.789} &\textbf{0.786} &0.684 &0.650 &0.667 \\
\textbf{2} &RRMS-B &0.786 &\textbf{0.789} &\textbf{0.787} &0.683 &0.652 &0.671 \\
\textbf{3} &SPMS-A &0.784 &\textbf{0.789} &\textbf{0.788} &0.682 &0.650 &0.669 \\
\textbf{4} &SPMS-B &0.708 &0.703 &0.697 &\textbf{0.737} &\textbf{0.712} &0.703 \\
\textbf{5} &PPMS-A &0.718 &0.716 &0.707 &\textbf{0.737} &0.709 &0.693 \\
\textbf{6} &PPMS-B &0.694 &0.699 &0.690 &0.702 &0.664 &\textbf{0.744}    \\ \hline
\end{tabular}}
\end{table}

Figure \ref{fig:define_example} shows the results for a test sample from one trial (RRMS-B) as segmented by the trial-conditioning model using CIN parameters from several different trials. The results clearly demonstrate unique segmentation styles across trials. One observation that can be made is that conditioning on the SPMS-B, PPMS-A and PPMS-B trials results in a noticeable relative over-segmentation on some of the larger lesions in all three cases. Using these three label styles also results in completely missing the lesion in the posterior part of the white matter tract located in the brain's midline. The similarity in the results across the SPMS-B, PPMS-B, and PPMS-A segmentation maps suggests that they form a potential annotation style subgroup. On the other hand, the RRMS-A and SPMS-A segmentation styles are similar to that of RRMS-B and therefore result in very accurate results according to the RRMS-B label. This suggests that these three styles make up another subgroup. While these groupings are noticeable in the qualitative results, we will show later in Section \ref{sec:cins} that these groupings are also revealed by our CIN parameter analysis.

In addition to qualitative results, the relationship between these sets of trials (SPMS-B, PPMS-B, PPMS-A) and (RRMS-A, SPMS-A, RRMS-B) is also demonstrated in the quantitative results presented in Table~\ref{tab:trialcond-f1}. This table shows the DICE performance on the test set for the conditioned model using the different trial styles. These results show that the best SPMS-B performance is obtained not only by the SPMS-B style, but also by the PPMS-A style. The PPMS-B style results in a SPMS-B performance that is intermediate between that of the SPMS-B/PPMS-A style and the RRMS-A/RRMS-B/SPMS-A style, suggesting that the PPMS-B style could be considered somewhat unique, depending on how the groupings are defined. The best RRMS-B performance is achieved by the RRMS-B, SPMS-A, and RRMS-A styles. The best PPMS-A performance is achieved by the SPMS-B style. These results further confirm a strong relationship between the trial labelling processes. The differences between the labelling processes are distinct but complex, and despite this, the scaling/shifting achieved by CIN parameters are able to capture these unique styles.

\subsection{Discovering Subgroupings in Aggregated Datasets with CIN Parameter Analysis}
\label{sec:cins}

We further explore and identify relationships between the trial styles by exploiting relationships in the learned CIN parameters (the trial-conditioned parameters in the CIN layer) from the trained trial-conditioned model discussed in Section~\ref{sec:trial_cond}. Recall that there are a set of learned CIN parameters for each trial in the training set. The CIN parameters consist of a scale and shift parameter, each of size [1, channels, 1, 1, 1], and there are two CIN layers per block in the network, rendering analysis challenging. In order to effectively evaluate relationships between high dimensional per-trial parameters, we take two approaches: 1) cosine similarity analysis, and 2) euclidean norm analysis.

In the cosine similarity analysis, we calculate the metric as described in Section~\ref{cin_method}. This analysis allows us to identify similarities between the directions of the CIN parameters, with $+1$ indicating the vectors are in the same direction in high dimensional space and $-1$ meaning they are in the opposite direction. This method allows us to gauge similarities between scales and shifts per layer, between each pair of trials. For the vector norm analysis, we plot the euclidean norm of the scale against the euclidean norm of the shift parameter on one plot for each layer, per-trial. This analysis results in 14 scatter plots where each point is the euclidean norm of the (scale, shift) for a different trial. This qualitative analysis allows for visualization of the different relationships considering both magnitudes. For the purpose of brevity, only one scatter plot per section of the network (encoder, center, decoder) is provided.

\begin{figure}[t]
\centering
  \centering
   \includegraphics[width=350pt]{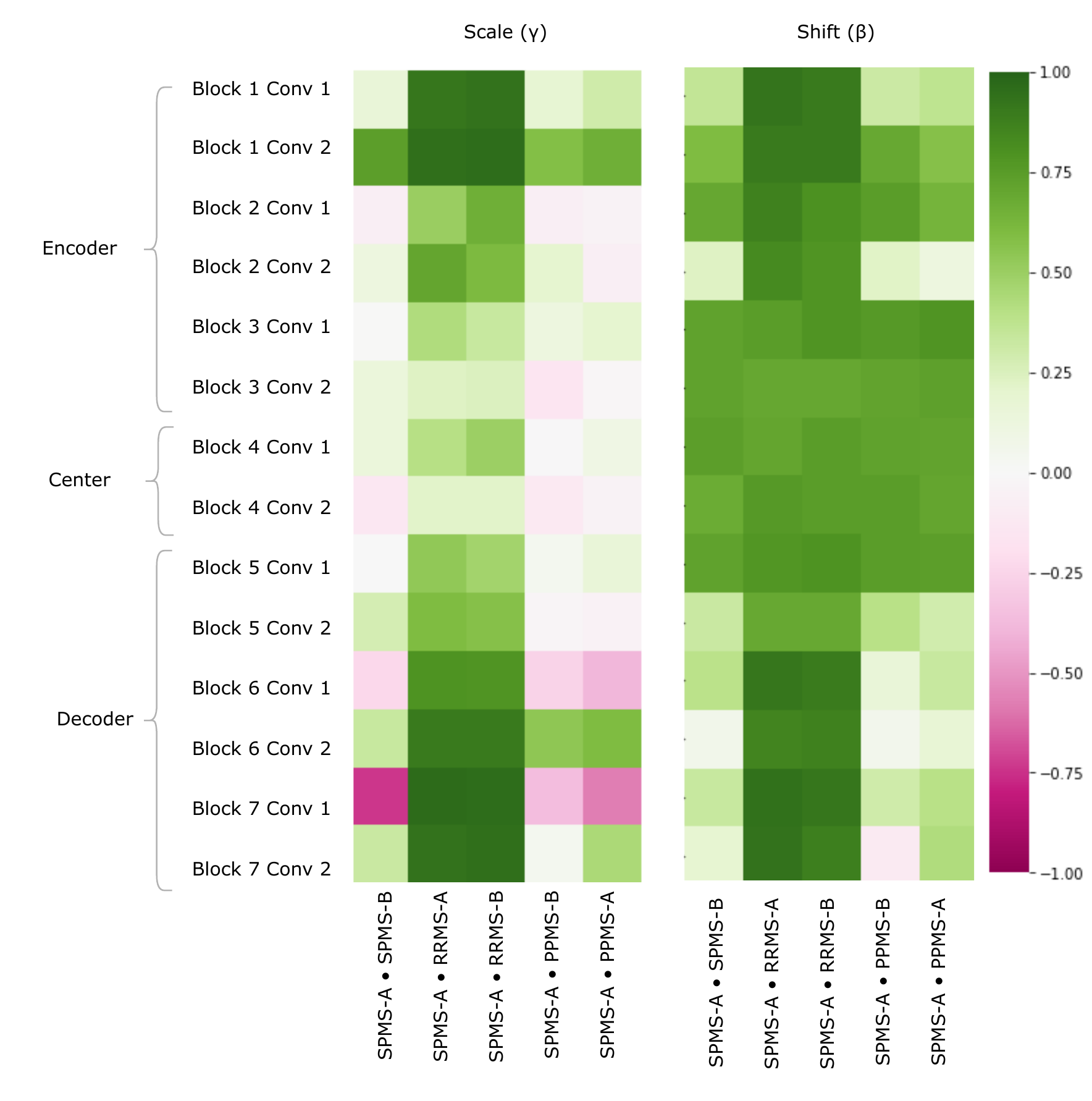}
  \caption{CIN parameter cosine similarity values between SPMS-A and all other trials for all CIN layers in the nnUNet.}
  \label{fig:SPMS-A_dots}
\end{figure}

\begin{figure}[t]
\centering
    \begin{subfigure}{0.77\textwidth}
        \includegraphics[width=\textwidth]{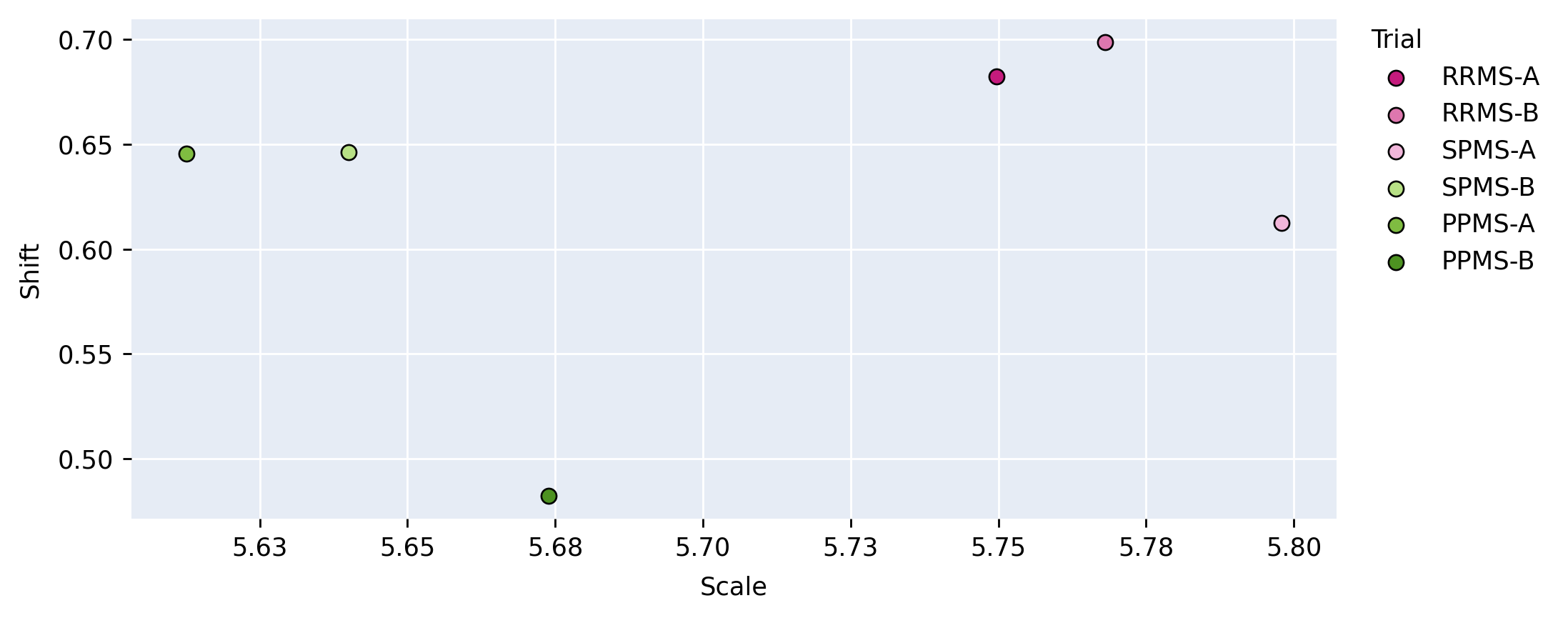}
        \caption{Encoder scatter plot example (Block 1 Convolution 2)}
        \label{fig:first}
    \end{subfigure}
    \hfill
    \begin{subfigure}{0.77\textwidth}
        \includegraphics[width=\textwidth]{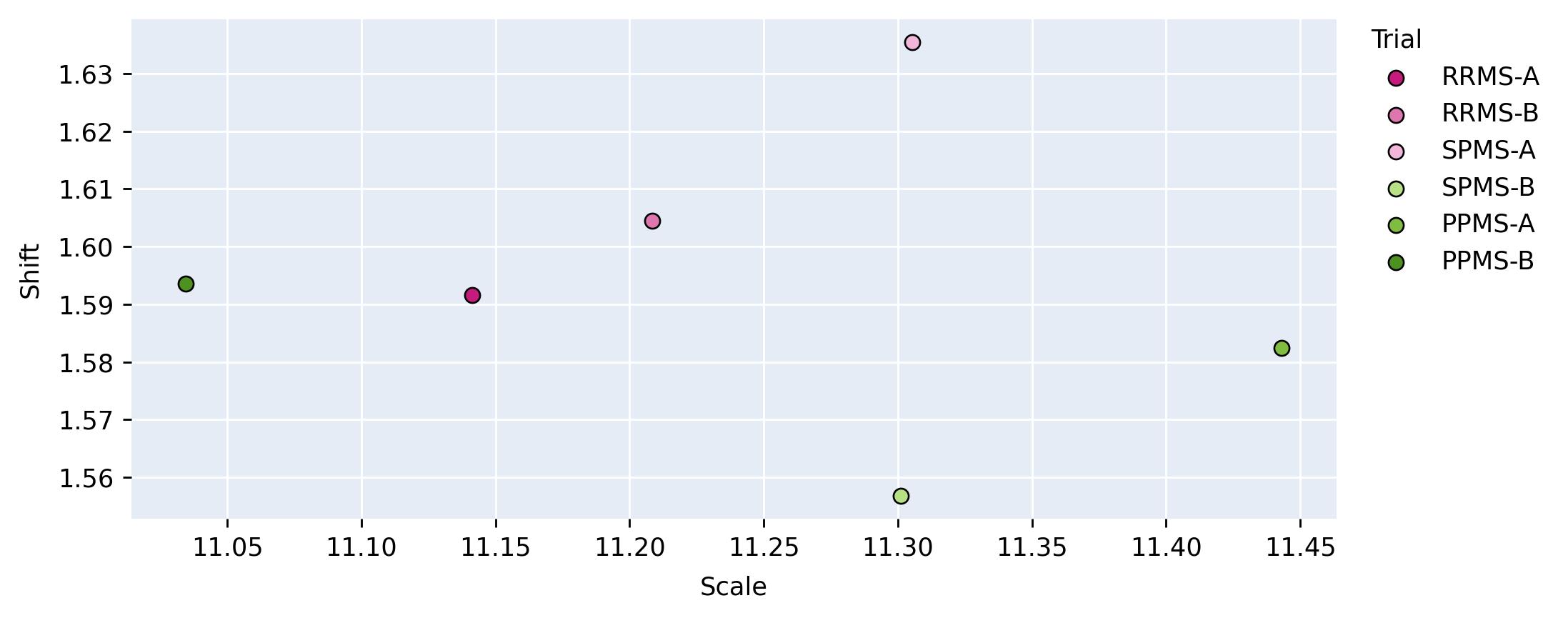}
        \caption{Center block scatter plot example (Block 4 Convolution 2).}
        \label{fig:second}
    \end{subfigure}
    \hfill
    \begin{subfigure}{0.77\textwidth}
        \includegraphics[width=\textwidth]{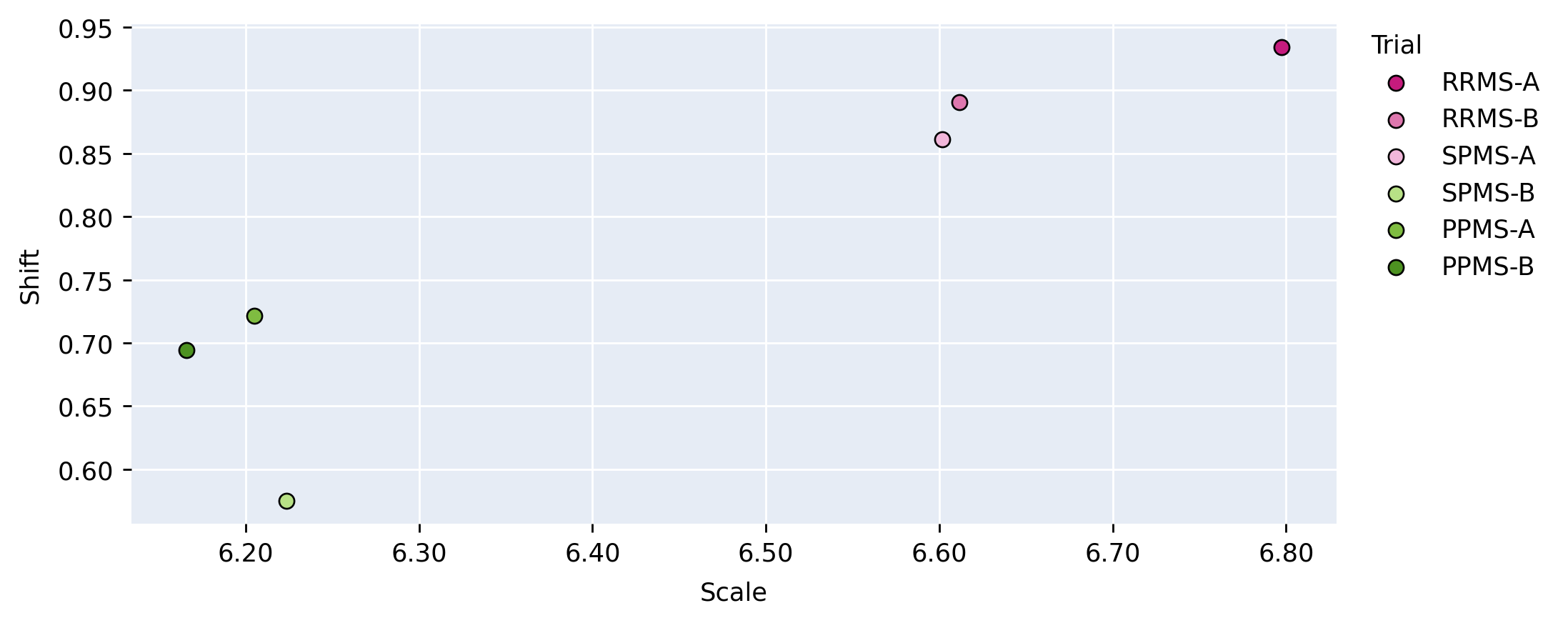}
        \caption{Decoder scatter plot example (Block 7 Convolution 2).}
        \label{fig:third}
    \end{subfigure}
    \caption{Scatter plots where each point shows the euclidean norm of scale and shift per trial for different layers. One example scatter plot is provided for each portion of the network (Encoder, Center, Decoder).}
      \label{fig:cin_scatter}
\end{figure}

Figure~\ref{fig:SPMS-A_dots} provides a visualization of the (pairwise) cosine similarity analysis for SPMS-A with respect to all other trials. This result reveals the subgroup trends noted in Section~\ref{sec:trial_cond}, where SPMS-A shows distinct similarity in annotation style with RRMS-A and RRMS-B. The analysis results for all other trials are shown in Appendix A. The same relationships are confirmed in all other trial comparisons, where two distinct style subgroups are discovered: (1) SPMS-B, PPMS-A, and PPMS-B, and (2) RRMS-A, RRMS-B, and SPMS-A. The main commonality between the groups is that the trials were labelled using different versions of the label generation pipeline, indicating that the label generation pipeline was the main contributor to annotation bias between these particular trials. This relationship is further shown in the scatter plots in Figure~\ref{fig:cin_scatter} where the two groups form visible clusters. Figure~\ref{fig:cin_scatter} shows proximity between the PPMS-A/PPMS-B/SPMS-B group with some distance from the cluster formed by the RRMS-A/RRMS-B/SPMS-A group. 

Other than the annotation style group identification, the cosine similarity analysis also uncovered other trends. Interestingly, the shifts are more similar across all trial comparisons than the scales, suggesting high discriminatory importance of the scale parameter in the CIN formula. The center layers were more similar across trials relative to the rest of the network layers, especially in the bias parameters. This could mean that there may not be a huge distinction between trials when they are encoded into a (high-level) latent space.

Lastly, the PPMS-B trial results in Figure \ref{fig:olympus_dots} shows very little strong relation or opposition to most of the trials. Although the similarity with PPMS-A and SPMS-B still noticeable, it is much less extreme than the similarity between just PPMS-A and SPMS-B. Figure \ref{fig:cin_scatter} also shows that the PPMS-B point strays further from PPMS-A and SPMS-B. Figure \ref{fig:cin_scatter} b) especially shows PPMS-B very astray from all other trial parameters. This trial's unique annotation could be due to a number of reasons, including unique changes or updates to the labelling or preprocessing pipeline, or a specific labelling protocol required for a particular study. These possible changes could all result in PPMS-B's unique CIN parameters. Although similar to SPMS-B and PPMS-A, these trends lead us to believe the PPMS-B trial could be considered its own group.

These identified groupings contradict some of the clinical similarities between trials, specifically the disease subtype. While one might think that naively pooling trials of the same disease type would be appropriate, our analysis shows the importance of considering other factors in the labelling process. In the case of this study, the likely cause of the differences between trials was a difference in the overall labelling process and in particular, the corresponding semi-automated labelling algorithm. The significant difference in the labelling process highlights the underlying ambiguity in MS lesion boarders. As previously noted, the non-lesional DAWM which can be found adjacent to both focal lesions and healthy tissue makes delineating lesion boarders or identifying smaller lesions somewhat subjective. The intensity distribution between DAWM and the focal lesions overlap, thus forcing both annotation software and human raters to decide on an arbitrary threshold to discretize what is in fact a continuous healthy-to-pathology transition \citep{seewann2009diffusely}. Making assumptions about annotation styles due to prior knowledge may therefore mislead researchers and result in unfair analyses or comparisons between different datasets.

\subsection{Using Group Discoveries from CIN Parameter Analysis}
The previous section outlined a strategy to explore annotation style relationships between trials by analysing similarities in the CIN parameters. The experimental results of the CIN parameter analysis indicated a strong relationship between the  RRMS-A, RRMS-B, and SPMS-A trials. SPMS-B and PPMS-A also showed high CIN parameter similarities, and PPMS-B was identified as unique. To validate the conclusions obtained from this method, another series of experiments using the identified groups are conducted in this section. Here, a single model is trained on each identified group (\textit{Group-pooling}) without any conditioning. Another model is trained with conditioning on the identified trial groupings. This \textit{Group-Conditioned} model was designed such that the trials within an identified group all share CIN parameters through the entire network during training. This results in one set of CIN parameters for SPMS-A, RRMS-A, and RRMS-B, and another set of parameters for SPMS-B and PPMS-A, and lastly a set for PPMS-B.

\begin{table}[h]
\caption{DICE performance on test sets for the group-based experiments.}
\begin{subtable}[h]{\textwidth}
\caption{Performance on test sets for Group-Pooling models, Group-Conditioning model, and the Trial-Conditioned model.}
\label{tab:grouped_v_baselines_f1}
\resizebox{\textwidth}{!}{%
\begin{tabular}{|l|ll|cccccc|}
\hline
\multirow{2}{*}{ } &
\multirow{2}{*}{\textbf{Model}} &
  \multirow{2}{*}{\textbf{\begin{tabular}[c]{@{}l@{}}Training \\ Set\end{tabular}}} &
  \multicolumn{6}{c|}{\textbf{Testing Set (DICE)}} \\
& &\textbf{} &\textbf{RRMS-A} &\textbf{RRMS-B} &\textbf{SPMS-A} &\textbf{SPMS-B} &\textbf{PPMS-A} &\textbf{PPMS-B} \\\hline
\multirow{2}{*}{\textbf{1}} &\multirow{3}{*}{\textbf{Group-Pooling}} &\multirow{3}{*}{\begin{tabular}[c]{@{}l@{}}Group 1: RRMS-A,\\ RRMS-B, SPMS-A\end{tabular}} &\multirow{3}{*}{\textbf{0.792}} &\multirow{3}{*}{\textbf{0.794}} &\multirow{3}{*}{\textbf{0.792}} &\multirow{3}{*}{0.677} &\multirow{3}{*}{0.640} &\multirow{3}{*}{0.644} \\
& & & & & & & & \\
\textbf{} & & & & & & & & \\
\multirow{2}{*}{\textbf{2}} &\multirow{2}{*}{\textbf{Group-Pooling}} &\multirow{2}{*}{\begin{tabular}[c]{@{}l@{}}Group 2: SPMS-B,\\ PPMS-A\end{tabular}} &\multirow{2}{*}{0.705} &\multirow{2}{*}{0.704} &\multirow{2}{*}{0.693} &\multirow{2}{*}{0.735} &\multirow{2}{*}{0.714} &\multirow{2}{*}{0.696} \\
& & & & & & & & \\
\textbf{3} &\textbf{Group-Pooling} &Group 3: PPMS-B &0.669 &0.671 &0.663 &0.669 &0.646 &0.742 \\
\textbf{4} &\textbf{Trial-Conditioned} &\textbf{All} &0.787 &0.789 &0.788 &\textbf{0.737} &0.709 &0.744 \\
\textbf{5} &\textbf{Group-Conditioned} &\textbf{All} &0.788 &0.786 &0.784 &\textbf{0.737} &\textbf{0.715} &\textbf{0.746} \\ \hline
\end{tabular}}
\end{subtable}
\newline
\vspace{1em}
\newline
\begin{subtable}[h]{\textwidth}
\caption{Performance on test sets for the Group-Conditioned model using all conditioning styles.}
\label{tab:grouped_swap_f1}
\resizebox{\textwidth}{!}{%
\begin{tabular}{|l|l|cccccc|}
\hline
\multirow{2}{*}{ } &
\multirow{2}{*}{\textbf{\begin{tabular}[c]{@{}l@{}}Conditioning \\ Style\end{tabular}}} &
  \multicolumn{6}{c|}{\textbf{Testing Performance (DICE)}} \\
& &\textbf{RRMS-A} &\textbf{RRMS-B} &\textbf{SPMS-A} &\textbf{SPMS-B} &\textbf{PPMS-A} &\textbf{PPMS-B} \\\hline
1 &{\begin{tabular}[c]{@{}l@{}}Group 1: RRMS-A,\\ RRMS-B,  SPMS-A\end{tabular}} &\textbf{0.788} &\textbf{0.786} &\textbf{0.784} &0.689 &0.658 &0.677 \\
2 &{\begin{tabular}[c]{@{}l@{}}Group 2: SPMS-B,\\ PPMS-A\end{tabular}} &0.709 &0.705 &0.694 &\textbf{0.737} &\textbf{0.715} &0.701 \\
3 &Group 3:PPMS-B &0.693 &0.695 &0.684 &0.7 &0.663 &\textbf{0.746} \\ \hline
\end{tabular}}
\end{subtable}
\end{table}

When naively pooling trials within the same identified group (group-pooling), the model does not suffer from the same performance degradation previously demonstrated in the all-trial naive-pooling experiments. In Table~\ref{tab:grouped_v_baselines_f1}, the resulting performance on the Group-Pooled model trained on SPMS-A, RRMS-A, RRMS-B group matches and slightly outperforms their single-trial baselines (see Table~\ref{tab:singletrial-f1}). The same is true for the SPMS-B and PPMS-A group-pooled model. These results confirm that the groups identified from the CIN parameter analysis do in fact have significant similarities between their annotation styles. The group-conditioned model also matches performance of the trial-conditioned model, further confirming the groups identified from the parameter analysis. These annotation style groups are especially beneficial for cases where each dataset is very small such that training on each independent dataset does not yield adequate performance in the desired annotation style.

Table~\ref{tab:grouped_swap_f1} also demonstrates that the grouped-conditioned model is successfully able to learn distinct styles even though several trials share CIN parameter sets. When using different annotation styles (conditioning on the ``wrong" group), DICE scores suffer up to 5\% performance degradation. This performance degradation further stresses the importance of understanding sources of annotation styles and their relationships. If researchers make assumptions about the sources of annotation styles, for example if they assume a disease-subtype cohort bias is the main cause of the annotation style, they can make incorrect groupings and combine datasets with incompatible annotation styles. For example, in this case, SPMS-B and PPMS-A are not the same disease subtype, but they have compatible annotation styles. Contrarily, PPMS-B is the same subtype as PPMS-A, but as shown in Table \ref{tab:grouped_swap_f1}, they have somewhat incompatible annotation styles resulting in the aforementioned performance degradation. 

These experiments also further argue the point that evaluating on subjective annotation styles is not truly fair. Although there is notable performance degradation when using the wrong conditioning/style, the model is clearly adequate as it is capable of performing well when it instructed to generate the desired annotation style. This finding emphasizes the importance of reconsidering how supervised models are evaluated in new settings.

\subsection{Accounting for Variable Annotation Styles in a Single-Source Dataset - Missing Small Lesions}
\label{sec:msl}
This set of experiments examines whether CIN is able to learn an isolated non-linear annotation style within a single dataset. In this experiment, we ensured that the source biases arose solely from different labelling protocols, while keeping all other factors, including disease stage, patient cohort, and time of collection, constant. To that end, only the SPMS-A dataset was used. For this experiment, the dataset consisted of 500 SPMS-A patient images. Half of the dataset was modified by changing the segmentation labels such that all small lesions (10 voxels or less) were set to the background class, essentially removing them from the label. This can be thought of as being equivalent to a labeling protocol that purposefully ignores small lesions for clinical purposes (Trial-MSL). The labels of the remaining half of the dataset were not modified in any way (Trial-Orig). These datasets were then each split into non-overlapping training (60\%), validation (20\%), and testing (20\%) sets. Similarly to the previous experiments, we train Single-Trial models, a Naive-Pooling model, and a CIN Trial-Conditioned model.

\begin{table}[t]
\scriptsize
\centering
\caption{Segmentation DICE and small lesion detection F1 scores shown on SPMS-A (Trial-Orig) test set using models trained on different combinations of the original dataset (Trial-Orig, 150 training patients) and the dataset with missing small lesions (Trial-MSL, 150 training patients).}
\resizebox{\textwidth}{!}{%
\begin{tabular}{|c|c|cc|cc|cc|}
\hline
\multirow{2}{*}{} & \multirow{2}{*}{\textbf{Model}}       & \multicolumn{2}{c|}{\textbf{Train Set}}                            & \multicolumn{2}{c|}{\textbf{Conditioned On}}                  & \multicolumn{2}{c|}{\textbf{Test Performance}}                      \\ \cline{3-8} 
                  &                                       & \multicolumn{1}{c|}{\textbf{Trial-Orig}} & \textbf{Trial-MSL}      & \multicolumn{1}{c|}{\textbf{Trial-Orig}} & \textbf{Trial-MSL} & \multicolumn{1}{c|}{\textbf{Sm Lesion F1}} & \textbf{Voxel Dice} \\ \hline
1                 & \textbf{Single-Trial}                    & \cmark                                   &                         & \multicolumn{2}{c|}{-}                                        & 0.795                                         & 0.844               \\
2                 & \textbf{Single-Trial}                    &                                          & \cmark                  & \multicolumn{2}{c|}{-}                                        & 0.419                                         & 0.837               \\
3                 & \textbf{Naive-Pooling}                    & \cmark                                   & \cmark                  & \multicolumn{2}{c|}{-}                                        & 0.790                                         & 0.797               \\ \hline
4                 & \multirow{2}{*}{\textbf{Trial-Conditioned}} & \multirow{2}{*}{\cmark}                  & \multirow{2}{*}{\cmark} & \cmark                                   &                    & 0.784                                         & 0.854               \\
5                 &                                       &                                          &                         &                                          & \cmark             & 0.496                                         & 0.850               \\ \hline
\end{tabular}}
\label{tab:m3-MSL-metrics}
\end{table}

Table \ref{tab:m3-MSL-metrics} shows the test results for this set of experiments on a non-modified Trial-C test set (Trial-Orig). We show detection results specific to small lesion detection in order to demonstrate CINs ability to learn an annotation style that ignores small lesions. Results show that the Single Trial-MSL model, which is trained on data with a significantly different annotation style than the test data, exhibits poor small lesion detection performance. The degradation in performance is expected as Trial-MSL has small lesions labeled as background, while the test set (Trial-Orig) has small lesions marked as lesions. The poor detection performance is not because the trained model is ``bad", as one may conclude from the results, but because the definition of ``ground truth" is different between the training set and the test set. The Naively-Pooled model, which is trained on both Trial-Orig and Trial-MSL was unable to learn the MSL style and suffered significantly in segmentation performance due to its inability to accommodate two conflicting annotation styles. The model trained using the CIN approach was able to adapt to the difference in annotation styles and exhibit good lesion-level detection and voxel-level segmentation performance when conditioned on the appropriate dataset. Looking at the Trial-conditioned model conditioned on Trial-MSL, we see that CIN is able to learn the Trial-MSL label bias quite effectively, and is able to ignore small lesions while maintaining voxel-level segmentation performance. This experiment shows the importance of providing context to the model regarding the desired annotation style in the test set. 

\begin{figure}[h]
\centering
  \centering
   \includegraphics[width=200pt]{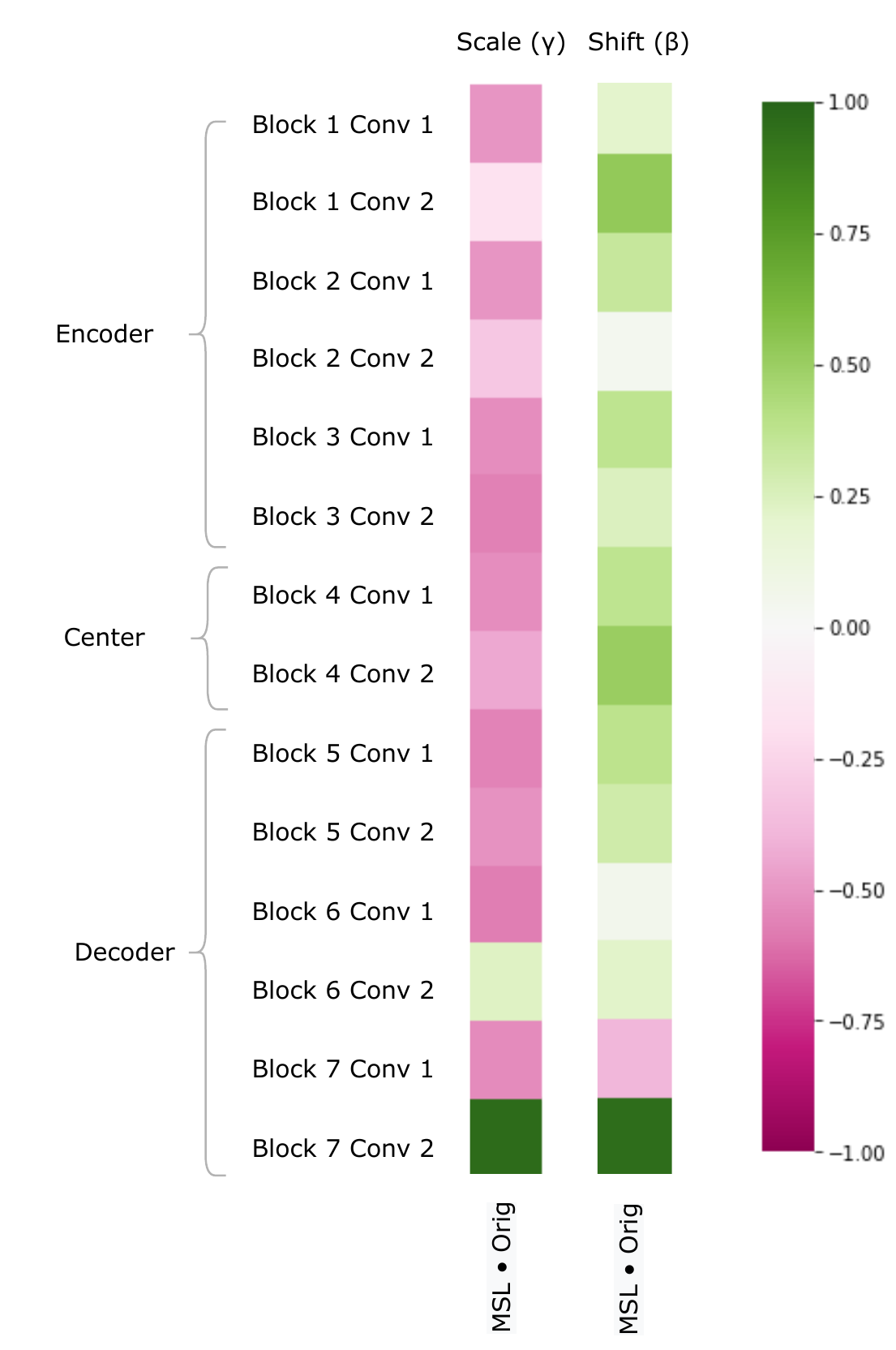}
  \caption{CIN parameter analysis for fine-tuning experiments.}
  \label{fig:finetune_dots}
\end{figure}

We can also study the impact of a pure annotation-based bias on the model by examining the CIN parameters as in Section~\ref{sec:cins}. Figure~\ref{fig:finetune_dots} clearly shows notable differences between the MSL and Orig trials, particularly in the scale parameter. These results demonstrate the impact annotation styles can have on network parameters, and further show CINs capacity to model complex differences in the label space. Interestingly, even though the bias was only in the annotations, there are very low cosine similarity values throughout the encoder. This is significant as it demonstrates that in order to model changes in the annotation style, the model makes use of capacity throughout the entire network. This also indicates that patients with different annotations styles (although potentially very similar in image appearance) can have completely different latent representations, thus highlighting the issue with some domain invariant methods that don't account for annotation style differences.

\subsection{Fine-tuning to a New Annotation Style with Limited Samples}
We now evaluate CINs ability to learn new annotation styles by mimicking a clinical situation where a new trial is introduced. In this experiment, we had two trials of ``known" bias, SPMS-B and RRMS-A, and we are provided with a few samples of a new ``unknown" trial, SPMS-A. We took two pre-trained models: one naively-pooling with SPMS-B and RRMS-A, and another CIN Trial-Conditioned model trained on only SPMS-B and RRMS-A. The affine parameters of the normalization layers were then fine-tuned using only 10 labeled samples from the SPMS-A dataset. Segmentations were done on a held-out test set from SPMS-A. In these experiments, we used 1000 SPMS-B samples and 1000 RRMS-A samples split into non-overlapping training (60\%), validation (20\%), and testing (20\%) sets. For SPMS-A, 10 samples were used for training, 100 were used for validation, and 100 for testing.

\begin{table}[h]
\scriptsize
\centering
\caption{DICE scores shown on the SPMS-A test set from the Naive-Pooling and Trial-Conditioned models trained on SPMS-B and RRMS-A. DICE scores are also shown for fine-tuned versions of those models, where the IN parameters were tuned using 10 SPMS-A samples.} 
\label{tab:finetune-metrics}
\begin{tabular}{|c|c|c|ccc|c|}
\hline
\multirow{2}{*}{} & \multirow{2}{*}{\textbf{Model}}       & \multirow{2}{*}{\textbf{\begin{tabular}[c]{@{}c@{}}Fine-Tuned\\ on SPMS-A\end{tabular}}} & \multicolumn{3}{c|}{\textbf{Conditioning Style}}                                                     & \multirow{2}{*}{\textbf{Test Performance}} \\ \cline{4-6}
                  &                                       &                                                                                           & \multicolumn{1}{c|}{\textbf{SPMS-B}} & \multicolumn{1}{c|}{\textbf{RRMS-A}} & \textbf{SPMS-A} &                                            \\ \hline
1                 & \multirow{2}{*}{\textbf{Naive-Pooling}}   &                                                                                           & \multicolumn{3}{c|}{-}                                                                           & 0.774                                      \\
2                 &                                       & \cmark                                                                                    & \multicolumn{3}{c|}{-}                                                                           & 0.819                                      \\ \hline
3                 & \multirow{3}{*}{\textbf{Trial-Conditioned}} &                                                                                           & \cmark                                &                                       &                  & 0.763                                      \\
4                 &                                       &                                                                                           &                                       & \cmark                                &                  & 0.806                                      \\
5                 &                                       & \cmark                                                                                    &                                       &                                       & \cmark           & 0.834                                      \\ \hline
\end{tabular}
\end{table}

Table \ref{tab:finetune-metrics} depicts the results from this set of experiments. The performance of the naively-pooled model improves when the instance normalization parameters of the model are fine-tuned compared to no fine-tuning. Furthermore, a CIN-pooled model shows good SPMS-A performance when conditioned on RRMS-A, which makes sense given the proven relationship between RRMS-A and SPMS-A. By fine-tuning the trial-specific CIN layer parameters of this model on 10 samples of SPMS-A, we are able to then condition the model on SPMS-A during test time. This leads to the highest performance improvement over all models. This experiment series shows how CIN can be leveraged in the clinic to quickly learn and accommodate the annotation style needed for the desired task. Additionally, it shows the need to consider annotation style in the evaluation or implementation of models, as poor results on a test set with a different annotation style can mislead researchers, who may otherwise assume that the performance drop stems from a difference in the image distribution (e.g. different scanners, etc.). As previously established, the RRMS-A style is compatible with the SPMS-A style, thus resulting in the relative good performance of the trial-conditioned model in RRMS-A-mode. Without considering subjective annotation styles, researchers may interpret this result as meaning the trial-conditioned model in RRMS-A-mode is better; however, it is simply because the RRMS-A style is more similar to the SPMS-A styled labels used in the test evaluation.

\begin{figure}[h]
\centering
  \centering
   \includegraphics[width=430pt]{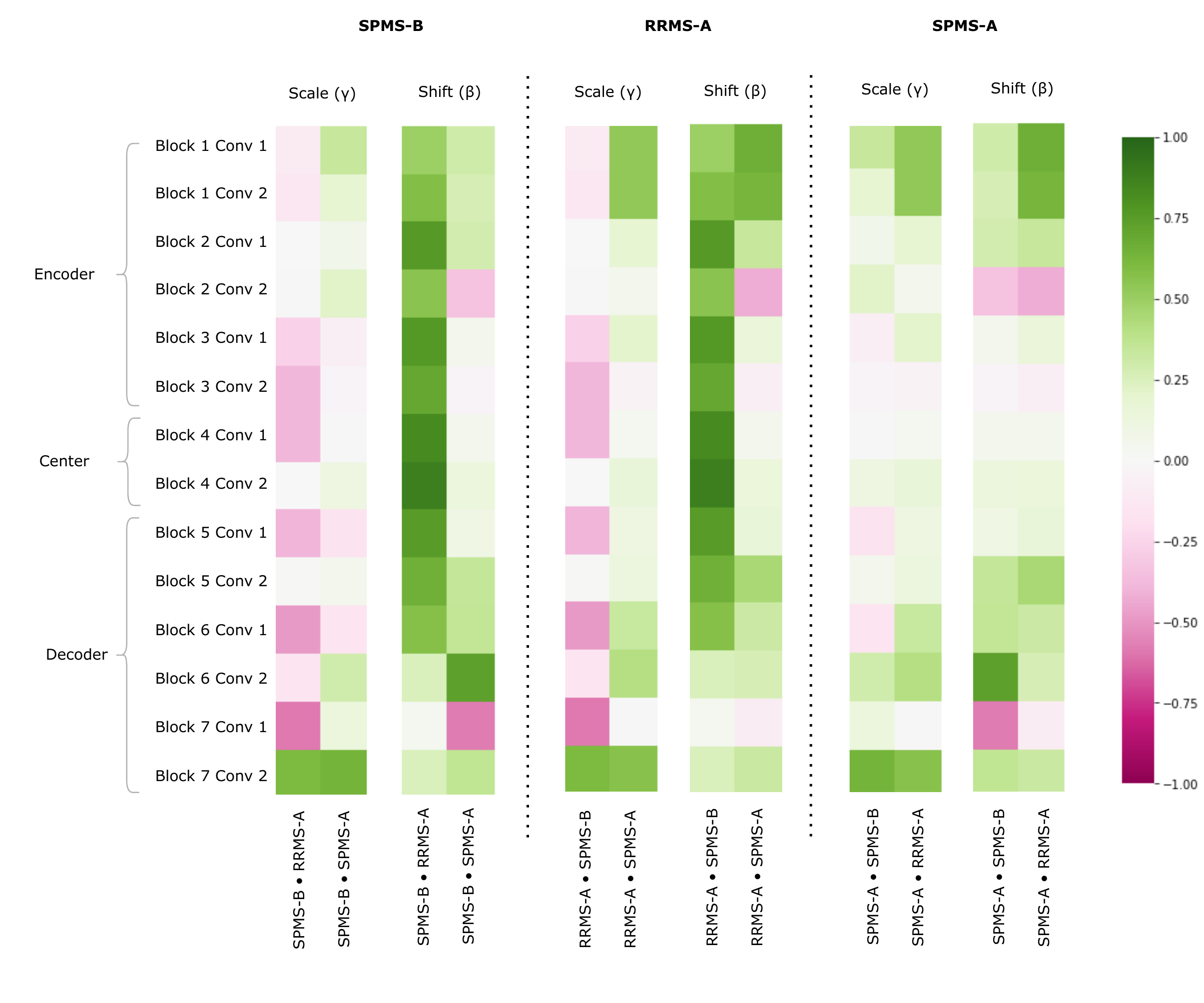}
  \caption{CIN cosine similarity parameter analysis for fine-tuning experiments.}
  \label{fig:finetune_dots}
\end{figure}

Figure \ref{fig:finetune_dots} shows the CIN Cosine Similarity analysis on the fine-tuning experiments~\footnote{Note that these experiments were done on an older version of our dataset, where the trials were preprocessed differently and not normalized as vigorously, which may account for trends in the CIN parameter analysis that do not corroborate the results presented in \ref{sec:cins}.} This analysis provides interesting insights about what CIN can learn in a few-shot fine-tuning problem. The fine-tuned SPMS-A parameters do not show particular relationship with either SPMS-B or RRMS-A. There are slightly more negative values in the SPMS-A $\cdot$ SPMS-B Scale parameters than there are in the SPMS-A $\cdot$ RRMS-A Scales, but this is not a clear cut relationship. This pattern does confirm some of the results in Table \ref{tab:finetune-metrics} as the RRMS-A CIN parameters yield better performance on the MAESTRO test set than do the SPMS-B. These figures may indicate that 10 samples is not enough for the algorithm to fully learn the annotation style, but the performance results do suggest that 10 is enough for the model to obtain reasonable performance. With only 10 samples, the algorithm is still able to perform better than it did than with only the SPMS-B and RRMS-A training samples. This information can come in the form of a few labelled samples in desired style, or in the form of clinically relevant factors. A hypothetical case for the latter would be if we knew that the RRMS-A style data was better at detection, and the SPMS-B style for lesion volume estimation, and the task of SPMS-A was lesion volume estimation, we would then make the call to use the style of the most relevant dataset, SPMS-B.

\subsection{Modeling Detection Biases Using Image Conditioning - A Synthetic Case Study}
In previous sections, we focused on learning and analyzing trial-specific biases. These biases were primarily in the label-space, and were primarily a consequence of known changes to the automatic segmentation method used across datasets. In this section, we examine whether the Image Conditioning mechanism proposed in Section~\ref{sec:imgcon_method} can model synthetic detection biases that can be derived from the image itself. To that end, we consider a scenario where the goal of a hypothetical study is to evaluate the effect of a particular drug on MS lesion activity in brain MRI. In this hypothetical scenario, we would be concerned with two key factors in order to evaluate the effect of treatment, principally: (1) Changes in T2 lesion volume between scans; and (2) The presence of gadolinium-enhancing lesions in brain MRI. Typically, significant increases in lesion volume \textit{or} the presence of gadolinium-enhancing lesions indicates treatment failure~\cite{Freedman2020-wf}. Practically speaking, gadolinium-enhancing lesions can be identified in the T1C sequence at a glance and do not require computing and comparing lesion volumes across multiple scans. In a resource constrained scenario, it is \textit{conceivable} that a rater, if gadolinium-enhancing lesions are clearly present, would not bother to make corrections to the T2 segmentation label produced by the automated segmentation algorithm, as the presence of a gadolinium-enhancing lesion is sufficient to indicate treatment failure. In this scenario, subjects that exhibit gadolinium-enhancing lesions would exhibit a different bias in the ``ground truth'' T2 lesion labels than those subjects that did not exhibit gadolinium-enhancing lesions. The patients belonging to the gadolinium-enhancing cohort would therefore require much more careful corrections to the T2 segmentation labels produced by the automated segmentation algorithm in order to determine whether the treatment was successful in suppressing the accumulation of additional T2 lesions. This example demonstrates a cohort-specific detection bias which, despite being correlated with the image, cannot be easily modeled given that dependencies are long range (Gadolinium enhancing lesion can appear anywhere in the brain).

Experimentally, to evaluate the ability of this mechanism to model synthetic detection biases, we induce a synthetic label bias to simulate the scenario we describe above. Specifically:
\begin{itemize}
  \item If a patient has gadolinium-enhancing lesions, we dilate the T2 lesion labels.
  \item If a patient does not have gadolinium-enhancing lesions, the T2 lesion labels are untouched.
\end{itemize}
Dilation is chosen since this emulates the scenario we describe above, where the rater would not bother to diligently delineate the edges of lesions in patients with gadolinium-enhancing lesions\footnote{Generally, raters correct the output of an automated method by removing false positives (rather than identifying false negatives).}. This synthetic dataset is generated using 1000 RRMS-A samples split into non-overlapping training (60\%), validation (20\%), and testing (20\%) sets. Overall, the proportion of patients that had gadolinium-enhancing lesions was approximately 34\%.

\begin{table}[t]
  \caption{DICE scores for T2 lesion segmentation shown on the RRMS-A test set. Results are stratified according to whether GAD lesions are present in the image or not, with each group having its own annotation bias.}
  \label{tab:image-conditioning-experiments}
  \scriptsize
  \centering
  \begin{tabular}{|c|c|c|cc|}
    \hline
    \multirow{2}{*}{} & \multirow{2}{*}{Model} & \multirow{2}{*}{\begin{tabular}[c]{@{}c@{}}Conditioned On Image\end{tabular}} & \multicolumn{2}{c|}{\begin{tabular}[c]{@{}c@{}}Test Performance\end{tabular}} \\ \cline{4-5} 
                  &                        &                                                                                 & \multicolumn{1}{c|}{w/o GAD}                             & w/ GAD                            \\ \hline
    1                 & Naive-Pooling          & -                                                                               & 0.659                                                    & 0.761                             \\ \hline
    2                 & Image-Conditioned      & \cmark                                                                            & \textbf{0.762}                                                    & \textbf{0.775}                             \\ \hline
  \end{tabular}
\end{table}

Table~\ref{tab:image-conditioning-experiments} depicts the results from this set of experiments. The performance of the naive-pooling model underperforms both single-trial baselines, with performance on the w/o GAD dataset suffering a dramatic performance hit. Clearly, the naive-pooling model is unable to learn the relationship between the presence of Gadolinium-enhancing lesions and the dilation bias while also maintaining performance on the un-altered labels for patient's without GAD. Indeed, the naive-pooling model's output appears to be biased toward the dilated dataset. On the other hand, the Image-Conditioned model is clearly able to learn the relationship between the presence of Gadolinium-enhancing lesions and the label bias as evidenced by the across the board performance gain relative to the naive-pooling model. In particular, the image-conditioned model outperforms the naive pooling model by 15.6\% on the w/o GAD dataset and by 1.8\% on the w/ GAD dataset. Notwithstanding, the image-conditioned model does underperform the single trial baseline models, suggesting that the image conditioning module is unable to detect GAD with perfect accuracy, which is not completely unexpected given that GAD lesions, in many cases, are as small as 3 voxels in size.

\section{Conclusions}
In this paper, we challenge traditional notions of generalization in the context of supervised medical image pathology segmentation models. We posit that poor generalization performance of automated methods across datasets may partially be a consequence of differing annotation styles, rather than solely the result of scanner/site differences as is often assumed. We point out that differing annotation styles can result from a number of factors, including the automated method used to generate the initial annotations (if any), the goal of the study (e.g. diagnosis, counting lesions, volumetric measurements), rater experience/education, and the instructions provided to raters, among others. Most importantly, we note that none of these factors are entirely random, but instead combine to constitute a systematic bias in the annotation process of each dataset. 

Given the myriad of factors that can influence the annotation style of any given dataset, we contend that annotation biases are unavoidable, necessitating a candid assessment of the inherent limits of generalizing to new datasets in a zero-shot manner. As such, we propose that embracing annotation biases, by modelling them, rather than ignoring or attempting to average them, poses a promising (and feasible) approach to deal with annotation differences across datasets. To do this, we propose using Conditioned Instance Normalization~\citep{CIN} to model the biases of individual datasets in the context of dataset aggregation, enabling a single model to generate multiple different annotation styles for a single test case. Next, we present an approach to identify relationships between different annotation styles, enabling the practitioner to strategically merge datasets with similar annotations in the low data regime. We also present a fine-tuning approach to adapt a fully trained model to a new dataset and corresponding annotation style with very little training data. Finally, we present an approach to model synthetic annotation biases that result from identifiable features in the input image affecting the annotation process. %


\acks{The authors are grateful to the International Progressive MS Alliance for supporting this work (grant number: PA-1412-02420), and to the companies who generously provided the clinical trial data that made it possible: Biogen, BioMS, MedDay, Novartis, Roche / Genentech, and Teva. Funding was also provided by the Canadian Institute for Advanced Research (CIFAR) Artificial Intelligence Chairs program, and a technology transfer grant from Mila - Quebec AI Institute. S.A.\ Tsaftaris acknowledges the support of Canon Medical and the Royal Academy of Engineering and the Research Chairs and Senior Research Fellowships scheme (grant RCSRF1819 /\ 8 /\ 25). Supplementary computational resources and technical support were provided by Calcul Québec, WestGrid, and Compute Canada. This work was also made possible by the end-to-end deep learning experimental pipeline developed in collaboration with our colleagues Eric Zimmerman and Kirill Vasilevski.}

%
\ethics{The work follows appropriate ethical standards in conducting research and writing the manuscript, following all applicable laws and regulations regarding treatment of animals or human subjects.}

\coi{Dr. Arnold reports consulting fees from Biogen, Celgene, Frequency Therapeutics, Genentech, Merck, Novartis, Race to Erase MS, Roche, and Sanofi-Aventis, Shionogi, Xfacto Communications, grants from Immunotec and Novartis, and an equity interest in NeuroRx. The remaining authors report no competing interests.}

\bibliography{sample}

\newpage
\appendix 
\section*{Appendix A.}
\label{sec:appendix_a}
\subsection*{Image Conditioning}
\begin{figure}[h]
\centering
  \centering
   \includegraphics[width=\textwidth]{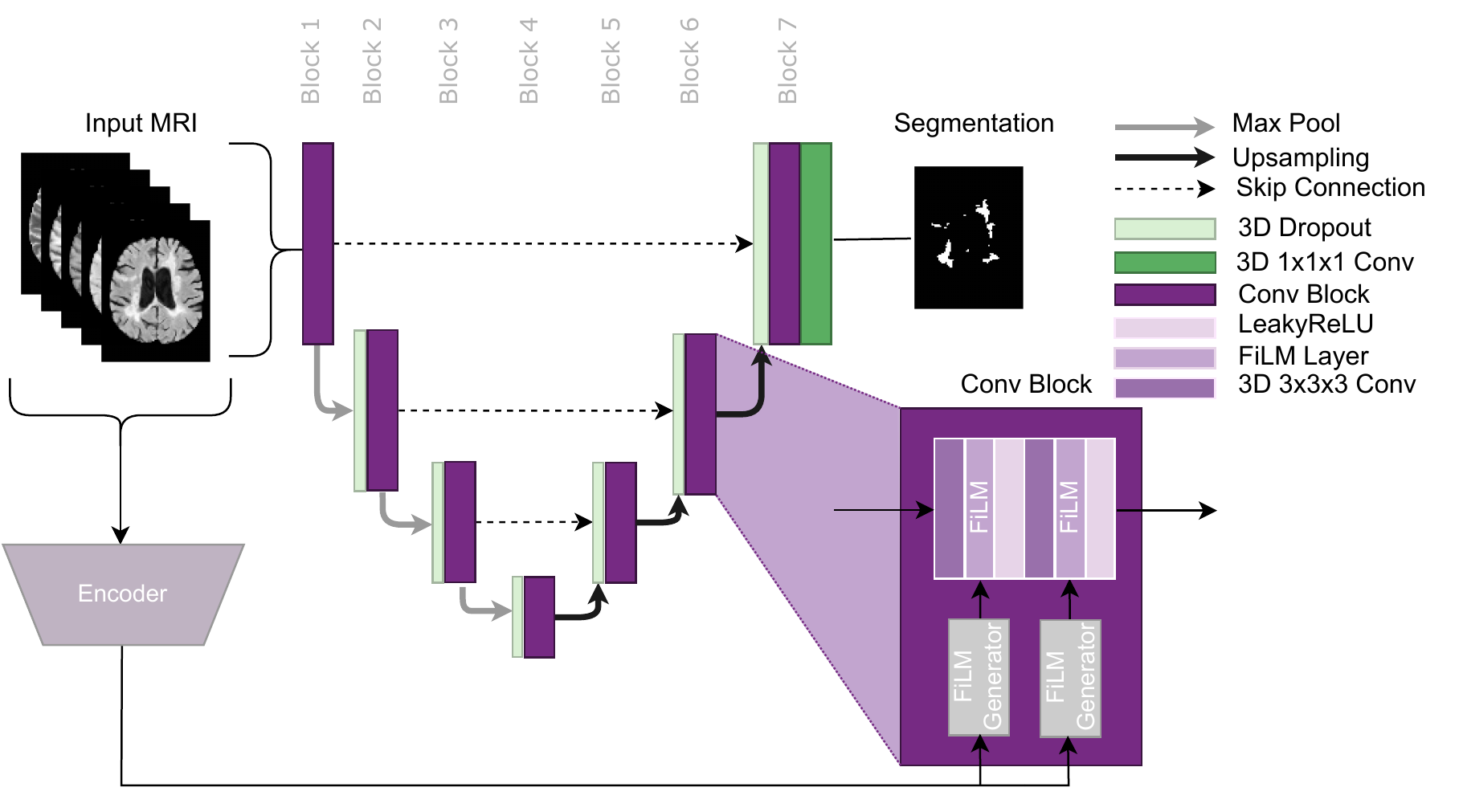}
  \caption{Left: Overview of modified nnUNet (\cite{nnUnet}) architecture used to segment MS T2 lesions with image conditioning. Right: Detail of an image conditioning conv block. It consists of a series of 3D 3x3x3 Convolution Layer, a FiLM layer and a linear FiLM generator, and a LeakyReLU activation layer.  }
  \label{fig:architecture_imcond}
\end{figure}
\subsection*{Hyperparameter Optimization}
Each network was optimized separately. Values for learning rate and schedulers, epochs, dropout, and data augmentation were varied for each network. For the single-trial baselines, the hyperparameters that resulted in the best validation performance on the training trial were chosen, and this model was tested against the other trial sets. For the multi-trial models, the hyperparameters that resulted in the best overall performance across all trials were used. The range for the hyperparameter search was as follows:
\begin{itemize}
    \item Epochs: 150-250
    \item Learning Rate: 1e-4 to 3e-4, with 1e-5 weight decay
    \item Scheduler:
    \begin{itemize}
        \item Epochs: [25, 50, 100, 150, 200] to [200]
        \item Gamma: 1/3 to 1/2
    \end{itemize}
\end{itemize}

\subsection*{Thresholding}

In this study, the threshold was selected on a per-trial basis to result in the highest F1. Specifically, for any given model throughout the paper, the results for a test set of a certain trial are presented at the threshold that yields the best F1 for that trial.

\subsection*{PR-AUC Results}

\begin{table}[h]
\caption{PR-AUC Performance on test sets: experiments on Single-Trial, Naive-Pooling, Trial-Conditioned models.}
\label{tab:cond_v_baselines_prauc}
\resizebox{\textwidth}{!}{%
\begin{tabular}{|l|llcccccc|}
\hline
\multirow{2}{*}{\textbf{ }} &
\multirow{2}{*}{\textbf{Model}} &
  \multirow{2}{*}{\textbf{\begin{tabular}[c]{@{}l@{}}Training \\ Set\end{tabular}}} &
  \multicolumn{6}{c|}{\textbf{Testing Set (metric PR-AUC)}} \\
& & &\textbf{RRMS-A} &\textbf{RRMS-B} &\textbf{SPMS-A} &\textbf{SPMS-B} &\textbf{PPMS-A} &\textbf{PPMS-B} \\\hline
\textbf{1} &\textbf{Single-Trial} &RRMS-A &0.875 &- &- &- &- &- \\
\textbf{2} &\textbf{Single-Trial} &RRMS-B &- &0.866 &- &- &- &- \\
\textbf{3} &\textbf{Single-Trial} &SPMS-A &- &- &0.873 &- &- &- \\
\textbf{4} &\textbf{Single-Trial} &SPMS-B &- &- &- &0.818 &- &- \\
\textbf{5} &\textbf{Single-Trial} &PPMS-A &- &- &- &- &0.773 &- \\
\textbf{6} &\textbf{Single-Trial} &PPMS-B &- &- &- &- &- &0.828 \\  \hline
\textbf{7} &\textbf{Naive-Pooling} &\textbf{All} &0.859 &0.843 &0.847 &0.809 &0.776 &0.822 \\
\textbf{8} &\textbf{Trial-Conditioned} &\textbf{All} &\textbf{0.879} &\textbf{0.878} &\textbf{0.879} &\textbf{0.826} &\textbf{0.786} &\textbf{0.829} \\ \hline
\end{tabular}}
\end{table}

\begin{table}[h]
\caption{PR-AUC performance on test sets for the Trial-Conditioned model with different conditioning styles.}
\label{tab:trialcond-prauc}
\resizebox{\textwidth}{!}{%
\begin{tabular}{|l|l|cccccc|}
\hline
\multirow{2}{*}{ } &
\multirow{2}{*}{\textbf{\begin{tabular}[c]{@{}l@{}}Conditioning \\ Style\end{tabular}}} & \multicolumn{6}{c}{\textbf{Testing Set (metric PR-AUC)}}      \\
                                       & &\textbf{RRMS-A} &\textbf{RRMS-B} &\textbf{SPMS-A} &\textbf{SPMS-B} &\textbf{PPMS-A} &\textbf{PPMS-B} \\\hline
\textbf{1} &RRMS-A &\textbf{0.879} &\textbf{0.878} &\textbf{0.878} &0.761 &0.717 &0.737 \\
\textbf{2} &RRMS-B &0.878 &\textbf{0.878} &\textbf{0.878} &0.760 &0.718 &0.742 \\
\textbf{3} &SPMS-A &0.876 &\textbf{0.878} &\textbf{0.879} &0.759 &0.717&0.740 \\
\textbf{4} &SPMS-B &0.796 &0.783 &0.781 &\textbf{0.826} &\textbf{0.791} &0.781 \\
\textbf{5} &PPMS-A &0.808 &0.801 &0.794 &\textbf{0.826} &0.786 &0.772 \\
\textbf{6} &PPMS-B &0.781 &0.783 &0.775 &0.782 &0.735 &\textbf{0.829}    \\ \hline
\end{tabular}}
\end{table}

\begin{table}[h]
\caption{ PR-AUC performance on test sets for Group-Pooling models, Group-Conditioning model, and the
Trial-Conditioned model.}
\label{tab:grouped_v_baselines_prauc}
\resizebox{\textwidth}{!}{%
\begin{tabular}{|l|ll|cccccc|}
\hline
\multirow{2}{*}{ } &
\multirow{2}{*}{\textbf{Model}} &
  \multirow{2}{*}{\textbf{\begin{tabular}[c]{@{}l@{}}Training \\ Set\end{tabular}}} &
  \multicolumn{6}{c|}{\textbf{Testing Set (metric PR AUC)}} \\
& &\textbf{} &\textbf{RRMS-A} &\textbf{RRMS-B} &\textbf{SPMS-A} &\textbf{SPMS-B} &\textbf{PPMS-A} &\textbf{PPMS-B} \\\hline
\textbf{1} &\textbf{Group-Pooling} & \begin{tabular}[c]{@{}l@{}}Group 1: RRMS-A,\\ RRMS-B, SPMS-A\end{tabular} &\textbf{0.883} &\textbf{0.883} &\textbf{0.883} &0.747 &0.696 &0.703 \\
\textbf{2} &\textbf{Group-Pooling} &\begin{tabular}[c]{@{}l@{}}Group 1: SPMS-B,\\ PPMS-A \end{tabular} &0.79 &0.784 &0.774 &0.823 &0.791 &0.772 \\
\textbf{3} &\textbf{Group-Pooling} &Group 3: PPMS-B &0.746 &0.744 &0.734 &0.746 &0.700 &0.828 \\
\textbf{4} &\textbf{Trial-Conditioned} &\textbf{All} &0.879 &0.878 &0.879 &\textbf{0.826} &0.786 &0.829 \\
\textbf{5} &\textbf{Group-Conditioned} &\textbf{All} &0.881 &0.876 &0.876 &\textbf{0.826} &\textbf{0.792} &\textbf{0.833} \\ \hline
\end{tabular}}
\end{table}

\begin{table}[h!]
\caption{PR-AUC performance on test sets for the Group-Conditioned model using all conditioning styles}
\label{tab:grouped_swap_prauc}
\resizebox{\textwidth}{!}{%
\begin{tabular}{|l|l|cccccc|}
\hline
\multirow{2}{*}{ } &
\multirow{2}{*}{\textbf{\begin{tabular}[c]{@{}l@{}}Conditioning \\ Style\end{tabular}}} &
  \multicolumn{6}{c|}{\textbf{Testing Performance (metric PR AUC)}} \\
& &\textbf{RRMS-A} &\textbf{RRMS-B} &\textbf{SPMS-A} &\textbf{SPMS-B} &\textbf{PPMS-A} &\textbf{PPMS-B} \\\hline
\textbf{1} &\begin{tabular}[c]{@{}l@{}}Group 1: RRMS-A,\\ RRMS-B, SPMS-A\end{tabular} &\textbf{0.881} &\textbf{0.876} &\textbf{0.876} &0.769 &0.728 &0.751 \\
\textbf{2} &\begin{tabular}[c]{@{}l@{}}Group 1: SPMS-B,\\ PPMS-A \end{tabular} &0.796 &0.784 &0.774 &\textbf{0.826} &\textbf{0.792} &0.781 \\
\textbf{3} &Group 3: PPMS-B &0.788 &0.779 &0.767 &0.78 &0.732 &0.833 \\ \hline
\end{tabular}}
\end{table}

\textcolor{white}{emptyempty}

\subsection*{Single-Trial Baseline Results}
\textcolor{white}{emptyempty}
\begin{table}[h]
\caption{PR-AUC performance on test sets: experiments on Single-trial models.}
\label{tab:singletrial-prauc}
\resizebox{\textwidth}{!}{%
\begin{tabular}{|l|ll|cccccc|}
\hline
\multirow{2}{*}{\textbf{ }} &
\multirow{2}{*}{\textbf{Model}} &
  \multirow{2}{*}{\textbf{\begin{tabular}[c]{@{}l@{}}Training \\ Set\end{tabular}}} &
  \multicolumn{6}{c|}{\textbf{Test Performance (PR-AUC)}} \\ 
& & &\textbf{RRMS-A} &\textbf{RRMS-B} &\textbf{SPMS-A} &\textbf{SPMS-B} &\textbf{PPMS-A} &\textbf{PPMS-B} \\\hline
\textbf{1} &\textbf{Single-Trial} &RRMS-A &\textbf{0.875} &0.865 &0.866 &0.750 &0.707 &0.704 \\
\textbf{2} &\textbf{Single-Trial} &RRMS-B &0.864 &0.866 &0.854 &0.746 &0.692 &0.706 \\
\textbf{3} &\textbf{Single-Trial} &SPMS-A &0.869 &\textbf{0.874} &\textbf{0.873} &0.751 &0.704 &0.705 \\
\textbf{4} &\textbf{Single-Trial} &SPMS-B &0.773 &0.764 &0.764 &\textbf{0.818} &\textbf{0.786} &0.766 \\
\textbf{5} &\textbf{Single-Trial} &PPMS-A &0.783 &0.782 &0.773 &\textbf{0.818} &0.773 &0.749 \\
\textbf{6} &\textbf{Single-Trial} &PPMS-B &0.746 &0.744 &0.734 &0.746 &0.700 &\textbf{0.828} \\ \hline
\end{tabular}}

\end{table}

\vspace{300pt}

\subsection*{Cosine Similarity Figures for All Trials}

\begin{figure}[h]
\centering
  \centering
   \includegraphics[width=\textwidth]{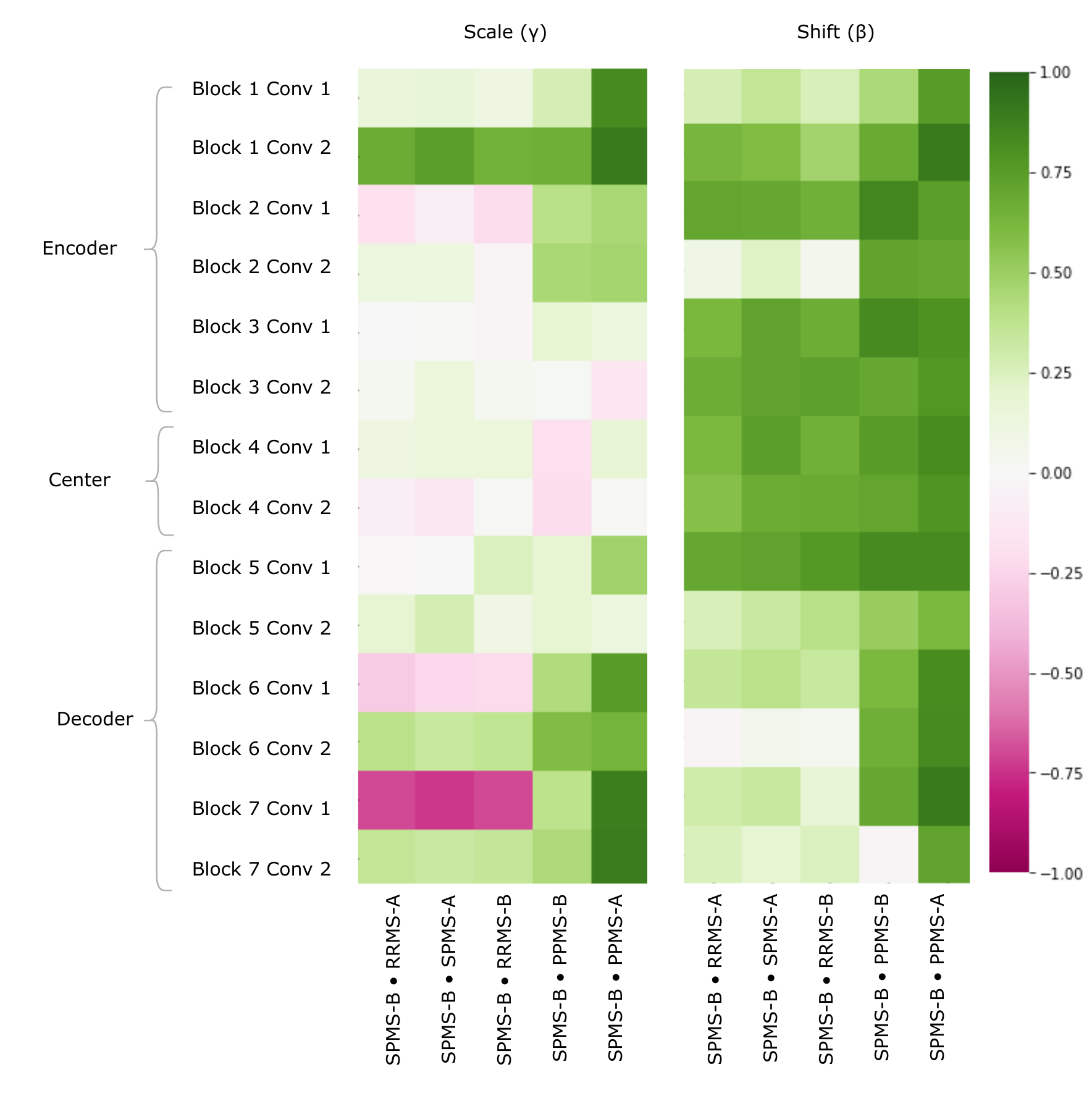}
  \caption{CIN parameter Cosine Similarity values between SPMS-B and all other trials for all CIN layers in the nnUNet.}
  \label{fig:SPMS-B_dots}
\end{figure}

\begin{figure}[h]
\centering
  \centering
   \includegraphics[width=\textwidth]{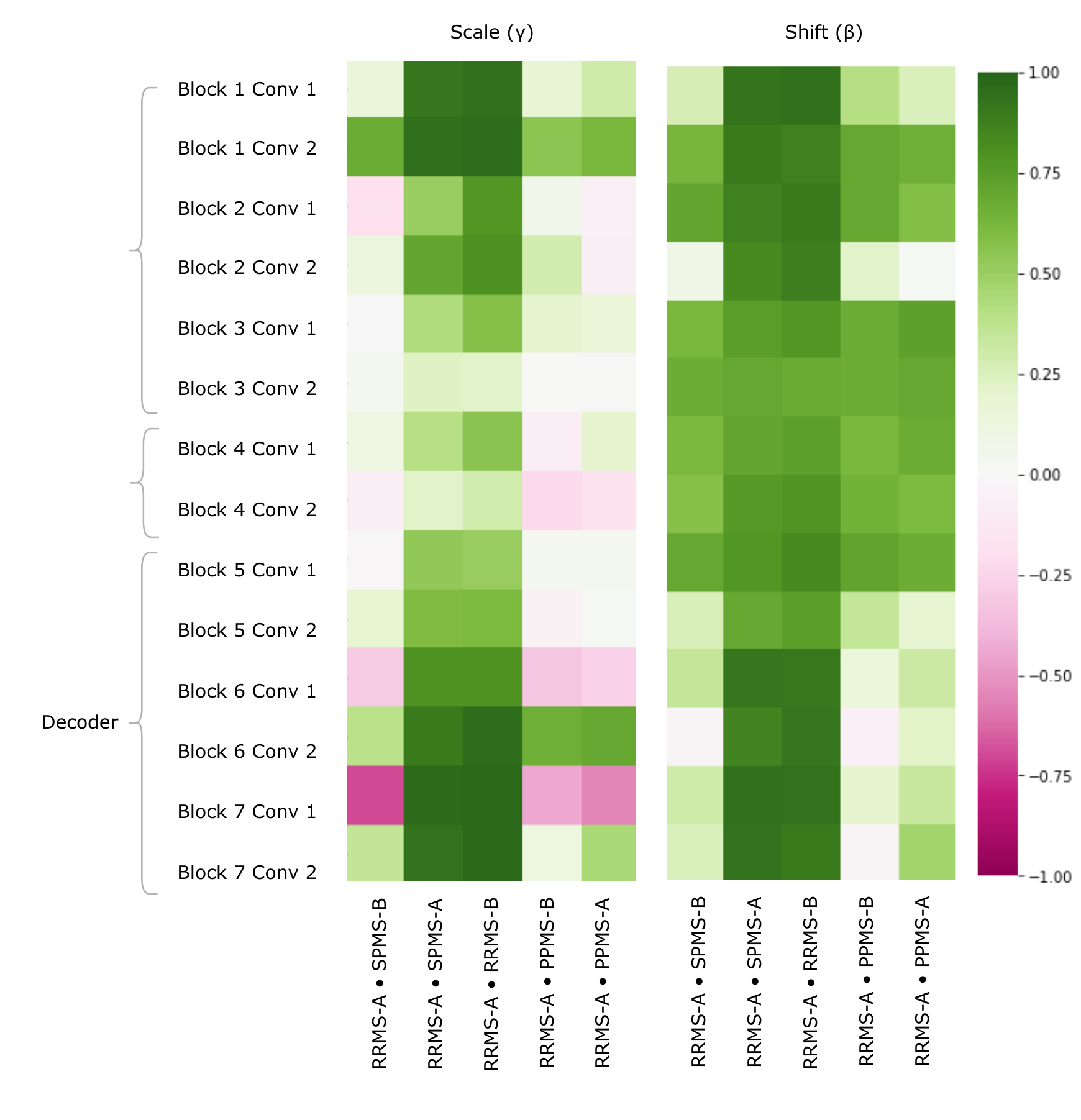}
  \caption{CIN parameter Cosine Similarity values between RRMS-A and all other trials for all CIN layers in the nnUNet.}
  \label{fig:bravo_dots}
\end{figure}

\begin{figure}[h]
\centering
  \centering
   \includegraphics[width=\textwidth]{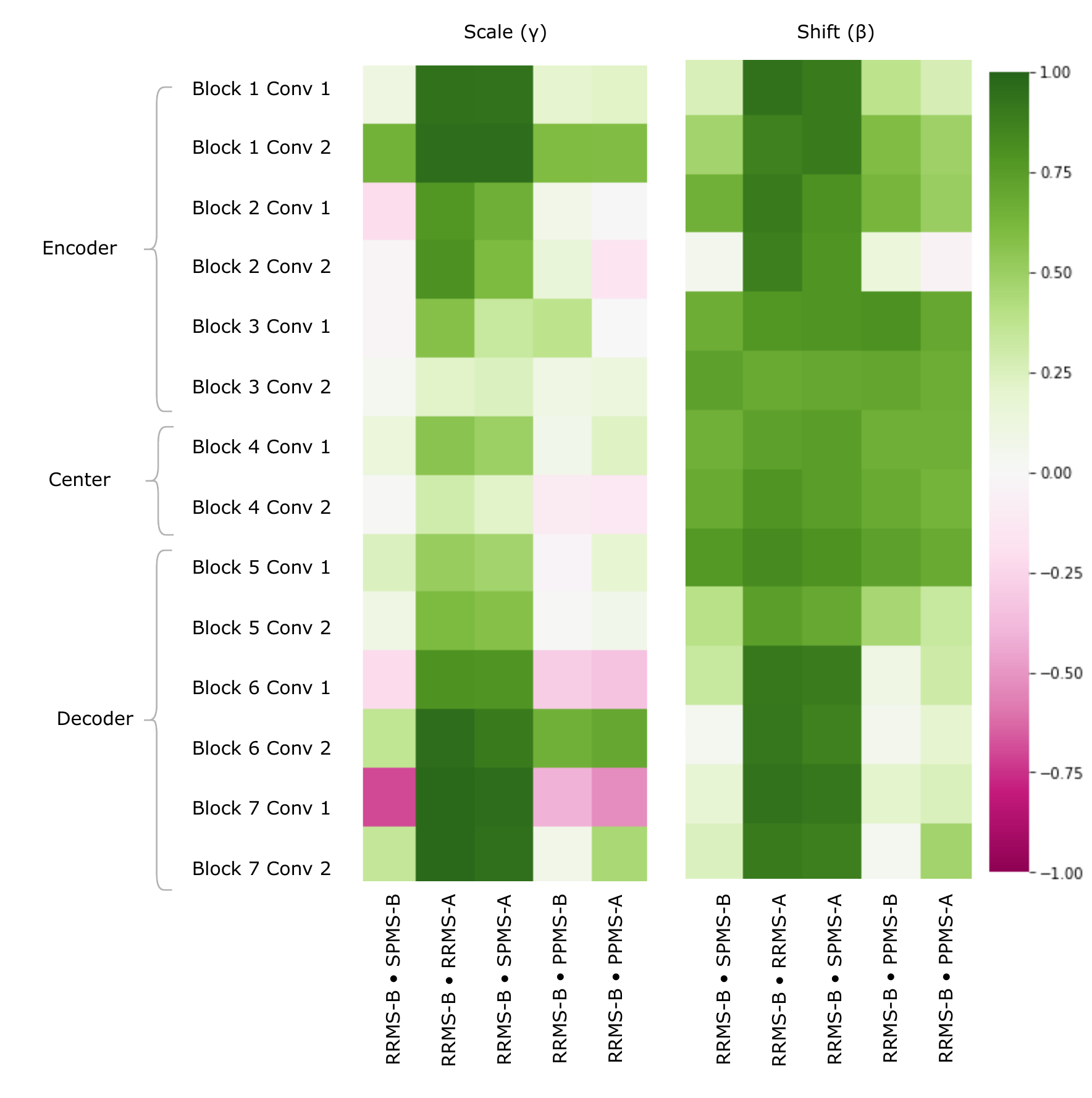}
  \caption{CIN parameter Cosine Similarity values between RRMS-B and all other trials for all CIN layers in the nnUNet.}
  \label{fig:define_dots}
\end{figure}

\begin{figure}[h]
\centering
  \centering
   \includegraphics[width=\textwidth]{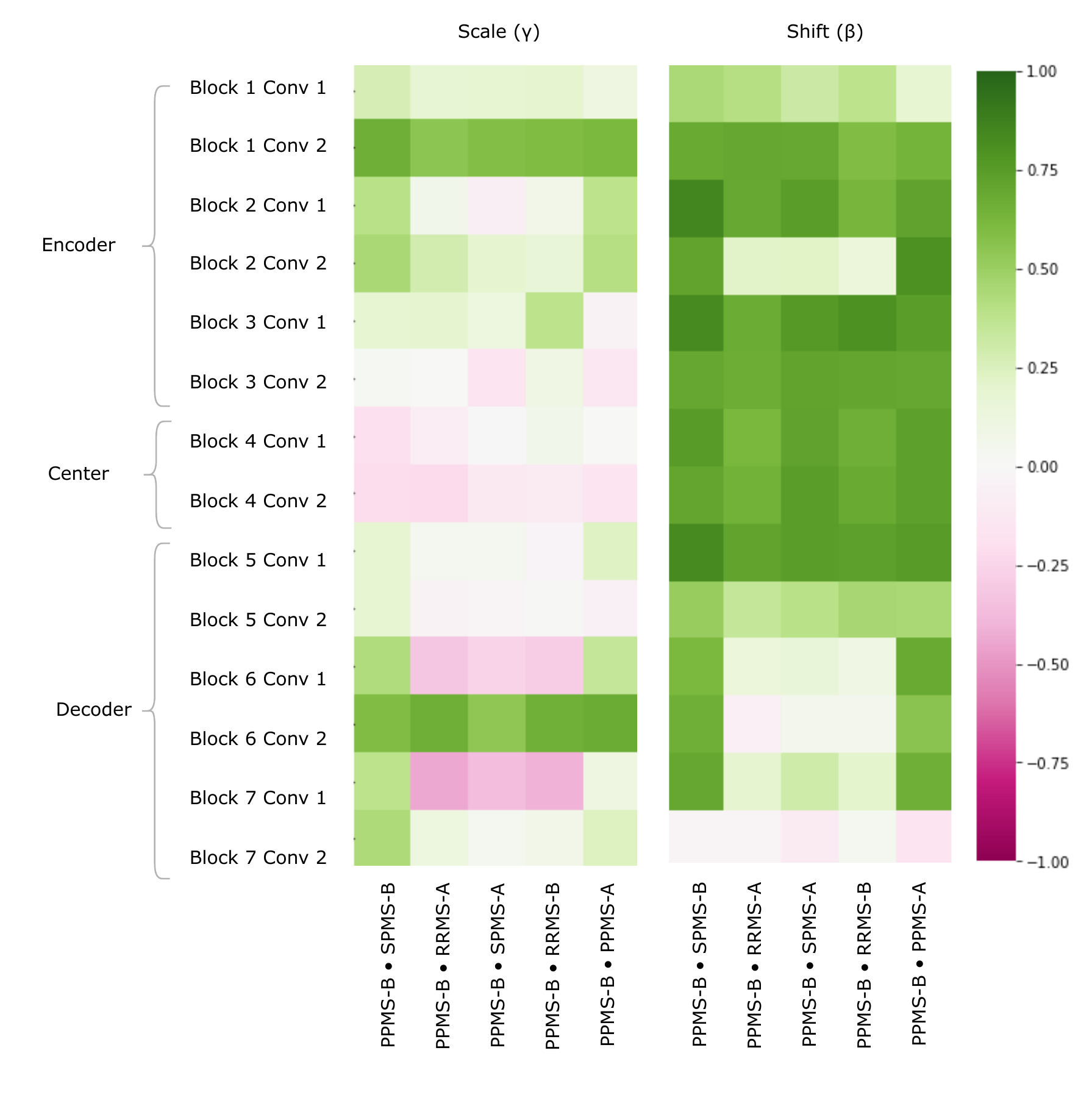}
  \caption{CIN parameter Cosine Similarity values between PPMS-B and all other trials for all CIN layers in the nnUNet.}
  \label{fig:olympus_dots}
\end{figure}

\begin{figure}[h]
\centering
  \centering
   \includegraphics[width=\textwidth]{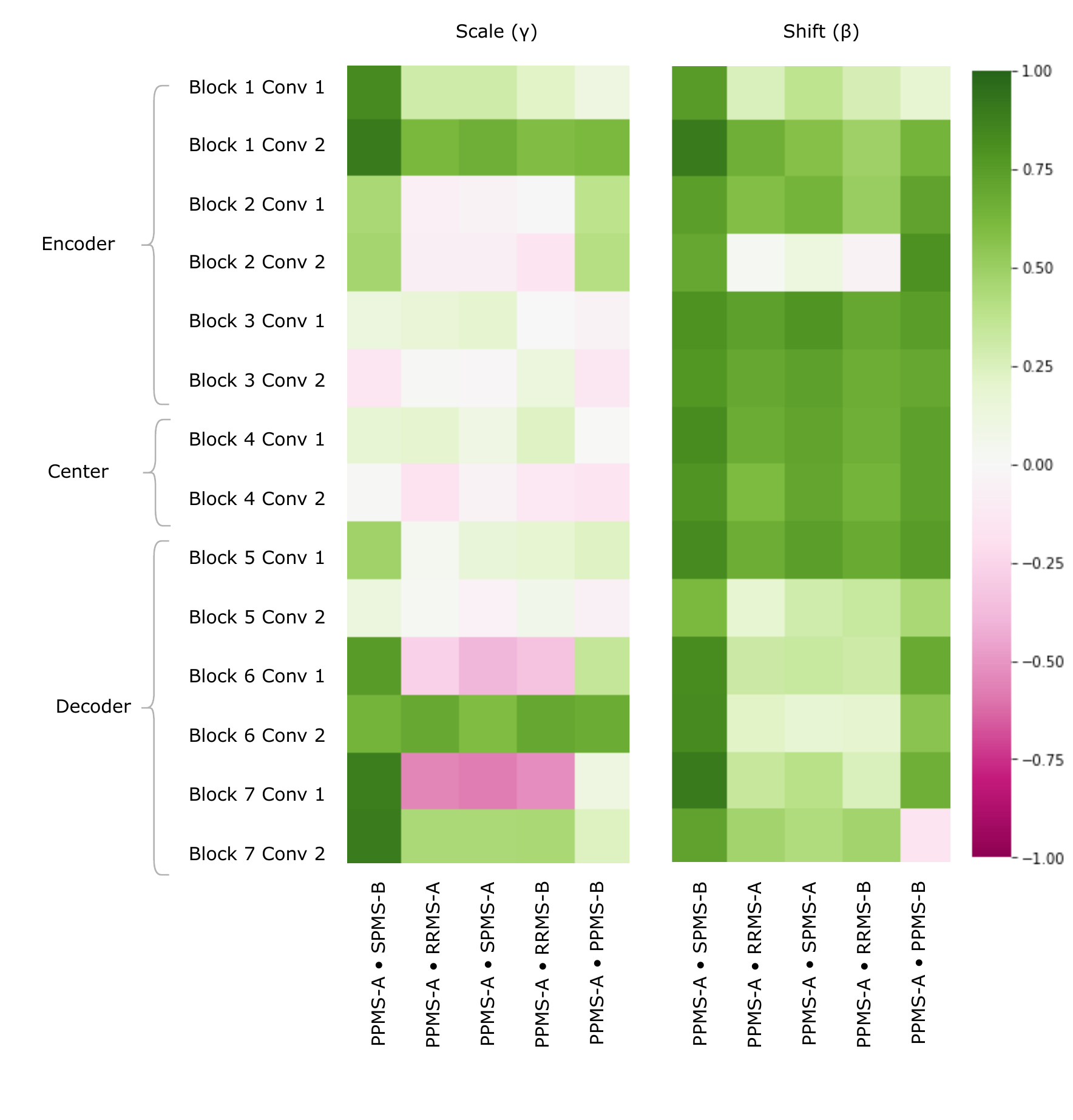}
  \caption{CIN parameter Cosine Similarity values between PPMS-A and all other trials for all CIN layers in the nnUNet.}
  \label{fig:oratorio_dots}
\end{figure}

\end{document}